\documentclass[sigconf,nonacm]{acmart}

\usepackage{afterpage}
\usepackage{multirow}
\usepackage{float}

\AtBeginDocument{%
  }

\begin{document}

\title{\raisebox{-0.25cm}{\includegraphics[width=1.5cm]{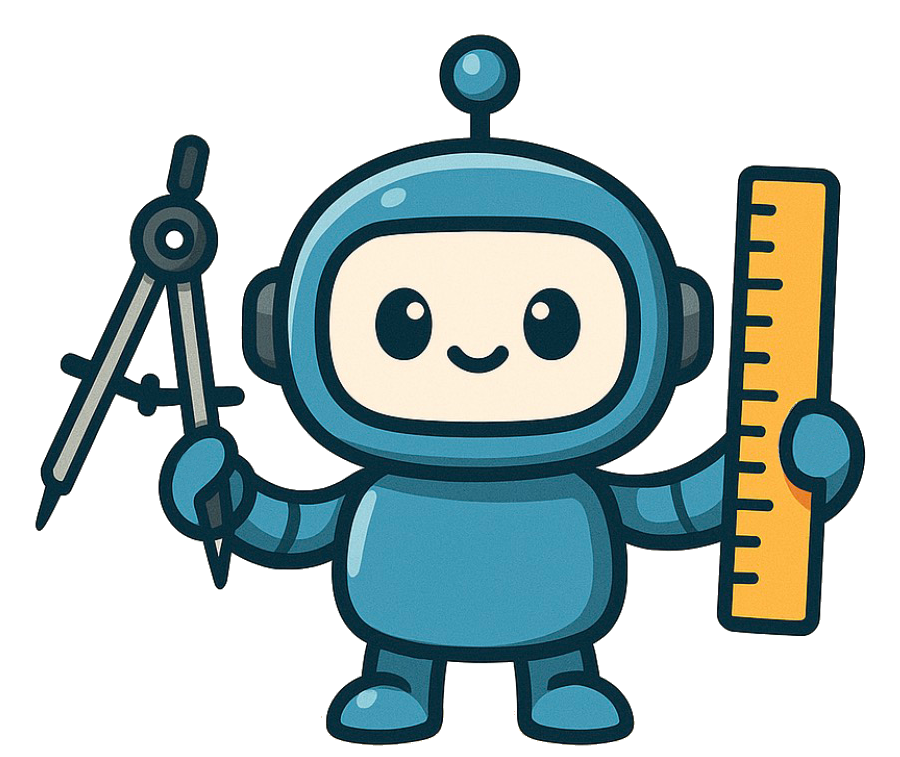}} GeoUni: A Unified Model for Generating Geometry Diagrams, Problems and Problem Solutions}

\author{Jo-Ku Cheng}
\authornote{Both authors contributed equally to this research.}
\email{chengruogu@stu.pku.edu.cn}
\affiliation{%
  \institution{School of Mathematical Sciences, Peking University}
  \city{Beijing 100871}
  \country{China}
}

\author{Zeren Zhang}
\authornotemark[1]
\email{eric_zhang@stu.pku.edu.cn}
\affiliation{%
  \institution{School of Mathematical Sciences, Peking University}
  \city{Beijing 100871}
  \country{China}
}

\author{Ran Chen}
\email{chenran@stu.pku.edu.cn}
\affiliation{%
  \institution{School of Mathematical Sciences, Peking University}
  \city{Beijing 100871}
  \country{China}
}

\author{Jingyang Deng}
\email{jingyang@stu.pku.edu.cn}
\affiliation{%
  \institution{School of Mathematical Sciences, Peking University}
  \city{Beijing 100871}
  \country{China}
}

\author{Ziran Qin}
\email{qinziran@sjtu.edu.cn}
\affiliation{%
  \institution{School of Electronic, Information and Electrical Engineering, Shanghai Jiao Tong University}
  \city{Shanghai 200240}
  \country{China}
}

\author{Jinwen Ma}
\authornote{Corresponding author.}
\email{jwma@math.pku.edu.cn}
\affiliation{%
  \institution{School of Mathematical Sciences, Peking University}
  \city{Beijing 100871}
  \country{China}
}

\begin{abstract}
We propose \textbf{GeoUni}\footnote{Our models are available at \url{https://github.com/chengruogu0915/GeoUni}.}, the first unified geometry expert model capable of generating problem solutions and diagrams within a single framework in a way that enables the creation of unique and individualized geometry problems. Traditionally, solving geometry problems and generating diagrams have been treated as separate tasks in machine learning, with no models successfully integrating both to support problem creation. However, we believe that mastery in geometry requires frictionless integration of all of these skills, from solving problems to visualizing geometric relationships, and finally, crafting tailored problems. Our extensive experiments demonstrate that GeoUni, with only 1.5B parameters, achieves performance comparable to larger models such as DeepSeek-R1 with 671B parameters in geometric reasoning tasks. GeoUni also excels in generating precise geometric diagrams, surpassing both text-to-image models and unified models, including the GPT-4o image generation. Most importantly, GeoUni is the only model capable of successfully generating textual problems with matching diagrams based on specific knowledge points, thus offering a wider range of capabilities that extend beyond current models. 
\end{abstract}

\begin{teaserfigure}
  \includegraphics[width=\textwidth]{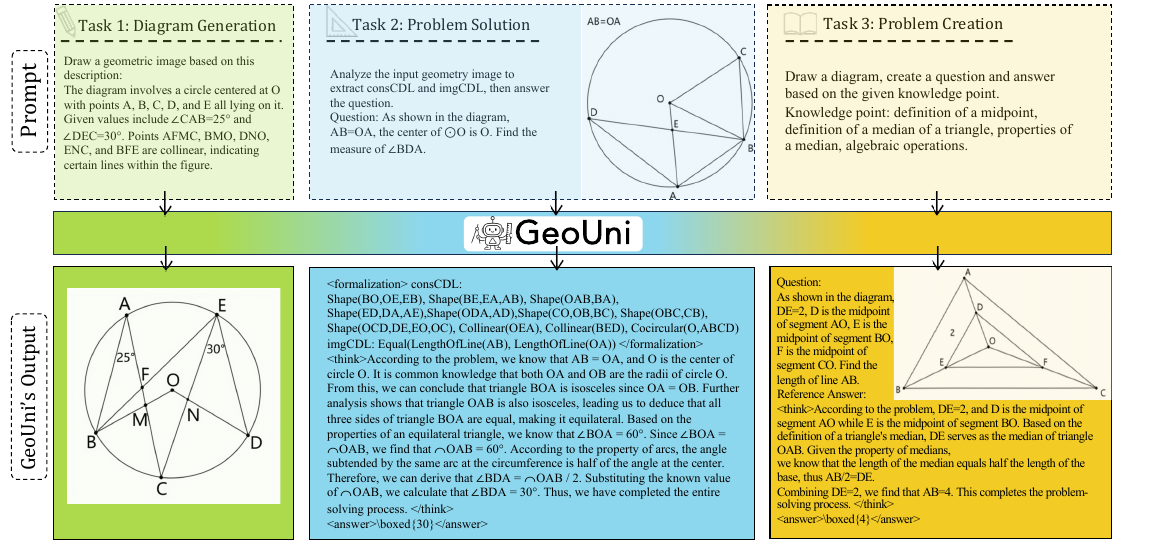}
  \caption{GeoUni can generate diagrams, solve problems and create new problems.}
  \Description{A figure shows the GeoUni feature.} 
  \label{fig:geouni}
\end{teaserfigure}


\maketitle
\section{Introduction}

\begin{quote}
    \textit{``If you want to master something, teach it.''} \\
    — Richard Feynman
\end{quote}

Mastery of geometry involves not only the ability to solve problems, but also the skills to analyze and visualize geometric relationships, as well as the ability to teach and tutor others by creating new problems that challenge then individually. Automated geometry problem solving and diagram generation have traditionally been separate fields. The former emphasizes mathematical reasoning, while the latter focuses on accurately representing topological relationships and generating proper alphanumerical and angle annotations on diagrams. Existing models typically address either problem solving or diagram generation, and they fall short in creating new problems that cater to the specific learning goals of a student, which is a crucial aspect of individualized learning in mathematics. For this reason, these models are limited to simulating student performance and incapable of effectively assuming a tutor role.

This limitation stems from the lack of a unified framework for problem creation that combines multi-modal geometry understanding with the ability to generate diagrams, corresponding problem textual descriptions, and reference answers simultaneously. When a model can solve problems, generate diagrams, and create new questions based on specific knowledge points, it transitions from a passive tool to an active educator. This transition enables the model to offer an individualized learning experience, much like a tutor who tailors questions to challenge the learner, fostering a more interactive and engaging educational environment.

Although a unified model must ultimately overcome the challenge of integrating multiple tasks, but its priority remains executing each separate task with precision. First, solving geometry problems requires both abstract textual reasoning and precise visual understanding. Some existing geometry solver, such as AlphaGeometry \cite{alphageo} and FGPS \cite{zhang2024fgeosss}, excel in geometric reasoning tasks but rely exclusively on textual inputs. There are also multi-modal models attempting to solve geometry problems \cite{gao2023gllava,zhang2025}. However, these models are limited to understanding diagrams and lack the ability to generate them.

Second, current diagram generation tools like GeoGebra \cite{geogebra} provide interactive graphical interfaces that rely heavily on manual user input through mouse interaction. Traditional text-to-image models, such as diffusion models \cite{Rombach2022} are primarily trained on natural images and thus struggle to generate accurate geometric diagrams. Even recent unified models like GPT-4o \cite{gpt4o}, despite significant advancements in general image generation, including textual content, still fall short in accurately plotting precise geometric diagrams.

%These limitations highlight the challenge of integrating geometry problem solving and diagram generation within a unified framework. 
To address these challenges, we propose \textbf{GeoUni}, the \textbf{first unified model} designed to integrate generating geometry problems, diagrams, and problem solutions seamlessly. As illustrated in Fig.~\ref{fig:geouni}, GeoUni demonstrates strong performance across text-to-diagram generation, geometric reasoning, and geometry problem creation tasks. Our model performance in geometric diagram generation surpasses existing models across various metrics. Additionally, GeoUni achieves geometric reasoning performance comparable to much larger models, accomplishing this with only 1.5B parameters across three datasets in both multiple choice and open-ended question modes. Furthermore, GeoUni demonstrates a unique capability in geometry problem generation that goes beyond the limitations of existing models. 

To facilitate better representation and tokenization of geometric diagrams, we propose \textbf{Geo-MAGVIT} designed to capture detailed geometric structures and reconstruct diagrams accurately. We observe that prior work, such as MagicGeo \cite{wang2025magicgeo}, evaluates diagram quality using the CLIP score. However, this metric is unsuitable for geometric diagrams, as CLIP is pre-trained on natural images and fails to capture the structural and symbolic characteristics. For a comprehensive evaluation of diagram quality, we introduce two new metrics: the \textbf{Geometry Semantic Matching Scores (GSMSs)}, which evaluates the alignment of geometry semantics, and the \textbf{Geometry Pixel Matching Score (GPMS)},  which assesses pixel-level fidelity. Additionally, we propose \textbf{Geo-Reasoning-Adapter}, which effectively leverages LoRA and GRPO to significantly enhance the model's reasoning ability for geometry problem solving without affecting its diagram generation capability.

The main contributions of this work are:
\begin{itemize}
\item We propose the first unified multi-modal geometry expert model, \textbf{GeoUni}, capable of solving geometry problems, generating precise geometric diagrams using both formal and natural language, and creating geometry problems based on knowledge points. All three tasks are supported in both English and Chinese.
\item  We propose \textbf{Geo-MAGVIT}, a module specifically designed for the tokenization of geometric diagrams. By introducing topo-structural awareness loss and text region loss, it significantly improves the precision of geometry structure and text reconstruction.
\item We innovatively combine GRPO and LoRA to train the \textbf{Geo-Reasoning-Adapter}, which effectively boosts geometric reasoning capability and seamlessly integrates into the unified model architecture.
\item We establish a novel diagram generation evaluation metrics, which includes the \textbf{Geometry Semantic Matching Scores (GSMSs)} and \textbf{Geometry Pixel Matching Score (GPMS)} to comprehensively evaluate the diagram generation task. 

\end{itemize}

\begin{figure*}[t] % 确保图片出现在第二页的顶部
  \centering
  \includegraphics[width=1.0\textwidth]{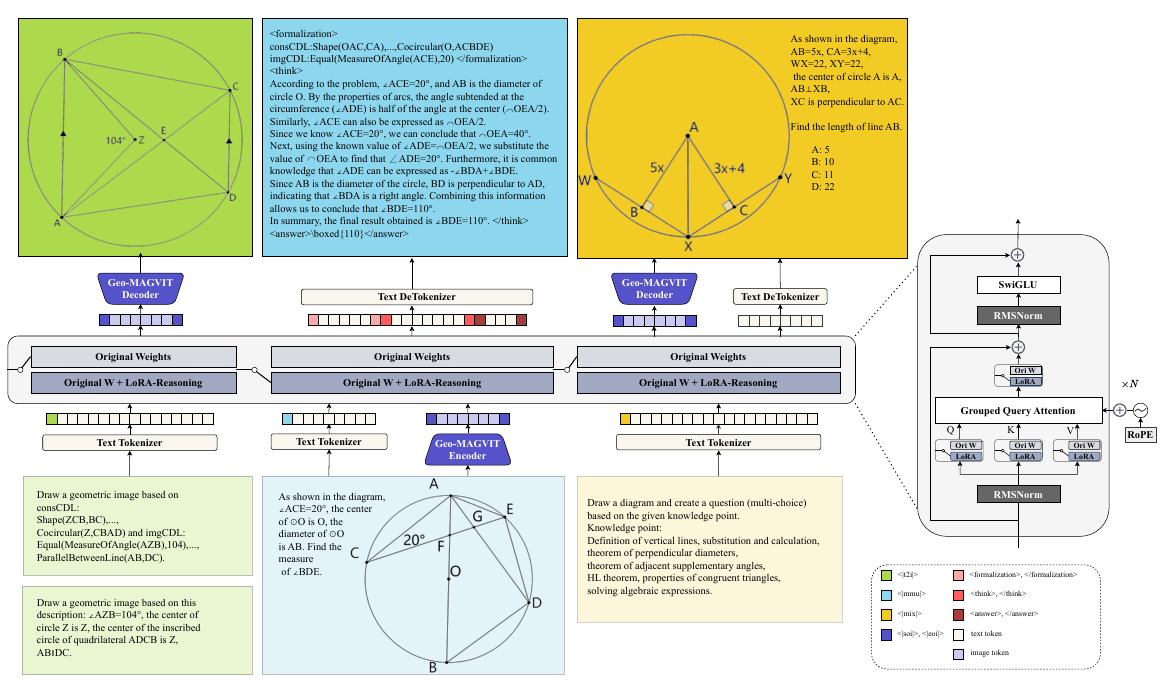}
  \caption{Overview of GeoUni}
  \Description{Overview of the proposed method.}
  \label{fig:overview}
\end{figure*}

\section{Related Work}
\subsection{Unified Model}

Multi-modal Large Language Models (MLLMs) \cite{liu2023, phi4mini2025, bai2023qwenvlversatilevisionlanguagemodel} are primarily designed to process images or videos as input and generate only text as output. This limitation has driven the development of unified models that integrate both multi-modal understanding and generation \cite{chen2025januspro, seedx, chameleon}. One of the first unified models, UNIFIED-IO \cite{lu2022unifiedio}, integrates text and image encoding into discrete tokens, enabling unified processing across multiple modalities. The Emu series \cite{sun2024emu, wang2024emu3} further unifies video, image, and text modeling within a next token prediction framework, while SEED-LLaMA \cite{ge2023seedllama} and Show-o \cite{xie2024showo} introduce techniques such as novel image tokenization and discrete diffusion modeling for improved performance. However, these models often struggle with geometric diagram generation, as diagrams present unique structural challenges not well addressed by standard image generation techniques.

\subsection{MLLM-based Geometry Problem Solver}

MLLM-based geometry problem solvers fall into two categories: those generating formal language programs requiring symbolic execution \cite{lu2021,zhang2023}, and those producing directly readable natural language answers.

The first category such as GeoX \cite{xia2025geox} introduces unimodal pre-training, geometry-language alignment, and end-to-end instruction tuning to train an MLLM capable of generating formal language reasoning steps. The second category, exemplified by G-LLaVA \cite{gao2023gllava}, follows the LLaVA training strategy, leveraging GPT-3.5 to construct the multi-modal geometry dataset, Geo170K. This dataset focuses on geometric cross-modal alignment and geometric instruction tuning to generate human-readable solutions. To address the mismatch between text descriptions and diagrams in Geo170K, DFE-GPS \cite{zhang2025} incorporates geometric formal language into diagram descriptions and creates a large-scale synthetic dataset SynthGeo228K to better train the Diagram Formalizer, enhancing the model’s ability to understand and generate accurate geometric representations. Despite these efforts, these models can understand diagrams, but still can not generate geometric diagrams.

\subsection{Automated Geometric Diagram Generation}

The Geometry Model Builder \cite{krueger2021} introduces the Geometry Model-Building Language (GMBL) to represent diagrams, treating the diagram creation process as a numerical optimization problem solved through gradient descent. Other approaches leverage natural language and LLMs to complete the process. GeoGPT4V \cite{geogpt4v} utilizes GPT-4 to generate Wolfram code, which is executed to produce the diagram. And MagicGeo \cite{wang2025magicgeo} prompts an LLM to formalize the diagram's description by encoding coordinate points and geometric constraints, which are then passed to a solver to find precise coordinate solutions. The LLM subsequently generates TikZ code to render the final diagram. However, all these models rely on generating formal language representations or code for rendering engines or solvers to construct diagrams, rather than adopting an end-to-end approach that directly generates diagrams from text.

\section{Preliminaries}
\subsection{Low-Rank Adaptation (LoRA)}
LoRA\cite{hu2021lora} is widely used for fine-tuning LLMs in various downstream tasks, as it preserves the performance of the base model while mitigating the issue of forgetting\cite{biderman2024}. The implementation of LoRA is straightforward: instead of updating the full weight matrix $W \in \mathbb{R}^{m \times n}$, it introduces two low-rank matrices, $A \in \mathbb{R}^{r \times n}$ and $B \in \mathbb{R}^{m \times r}$, where $r \ll \min(m, n)$. After training the low-rank matrices $A$ and $B$, the target weight is determined by the following expression:

\begin{equation}
W_{target}= W_{base}+ \Delta W = W_{base} + BA.
\end{equation}

\subsection{Group Relative Policy Optimization(GRPO)}
% 介绍一下背景知识
GRPO \cite{shao2024deepseekmath, deepseekr1} reduces the training costs of reinforcement learning (RL) by eliminating the need for a value model in the training loop. It utilizes the sampled outputs $\{o_1, o_2, \dots, o_G\}$ from the policy model to compute the corresponding rewards $\{r_1, r_2, \dots, r_G\}$, which are then used to compute the group normalized score as a relative advantage estimate:
\begin{equation}
\hat{A}_{i, t} = 
\frac{r_i - \text{mean}(\{r_1, r_2, \dots, r_G\})}{\text{std}(\{r_1, r_2, \dots, r_G\})}. 
\end{equation}

Then the policy model is optimized using the following objective:
\begin{equation}
\begin{aligned}
&\mathcal{J}_{GRPO}(\theta) = \mathbb{E}\left[q \sim P(Q), \{o_i\}_{i=1}^{G} \sim \pi_{\theta_{\text{old}}} (O|q) \right]  \\
& \frac{1}{G} \sum_{i=1}^{G} \frac{1}{|o_i|} \sum_{t=1}^{|o_i|}  
\left[ \frac{\pi_{\theta} (o_{i,t} | q, o_{i,<t})} {\left[\pi_{\theta_{\text{old}}} (o_{i,t} | q, o_{i,<t})\right]_{\text{no grad}}} \hat{A}_{i,t} - \beta \mathbb{D}_{KL} (\pi_{\theta} \| \pi_{\text{ref}})\right].
\end{aligned}
\end{equation}

\section{Methodology}

\subsection{Overview}

Our model, GeoUni, needs to address several challenges. First, because the unified model's vision tokenizer is trained on general images, it faces the same issues identified by \cite{zhang2025, xia2025geox}. This causes it to be ineffective in tokenizing geometric diagrams, and limits its ability to accurately reconstruct and generate them. Second, effectively integrating the three tasks into a unified training framework remains a non-trivial challenge. Finally, another key difficulty is to enhance the model's reasoning capabilities without compromising its diagram generation ability. To address these issues, the training pipeline is organized into three stages, each with its own focus:
%  as illustrate in Fig.~\ref{fig:overview}

\begin{itemize}
    \item \textbf{Diagram Tokenization Pretraining.} We propose \textbf{Geo-MAGVIT} to improve the tokenization of geometric diagrams. Building on MAGVIT\cite{luo2025openmagvit2}, we introduce geometric topo-structural awareness loss and text region loss to better reconstruct the topological structure and the text within the diagrams.
    
    \item \textbf{Multi-Task Instruction Tuning.} To achieve the geometry expert unified model, we propose the \textbf{Diagram Formalization Unified Prompting} method in multi-task instruction tuning for text-to-diagram generation, problem solving, and problem generation, achieving next-token prediction training. This training phase equips GeoUni with the capability to accurately generate geometric diagrams, solve basic geometry problems, and generate problems based on knowledge points.
    
    \item \textbf{Reasoning Enhancement.} We combine LoRA and GRPO to train the \textbf{Geo-Reasoning-Adapter}, which significantly improves the model's geometric reasoning ability while preserving its precise geometric diagram generation capability.

\end{itemize}

\subsection{Diagram Tokenization Pretraining}

Following MAGVIT \cite{luo2025openmagvit2}, we pre-train the Geo-MAGVIT on a geometry dataset consisting of approximately 200K diagrams.

\begin{figure}[htbp]
  \centering
  \includegraphics[width=1.0\linewidth]{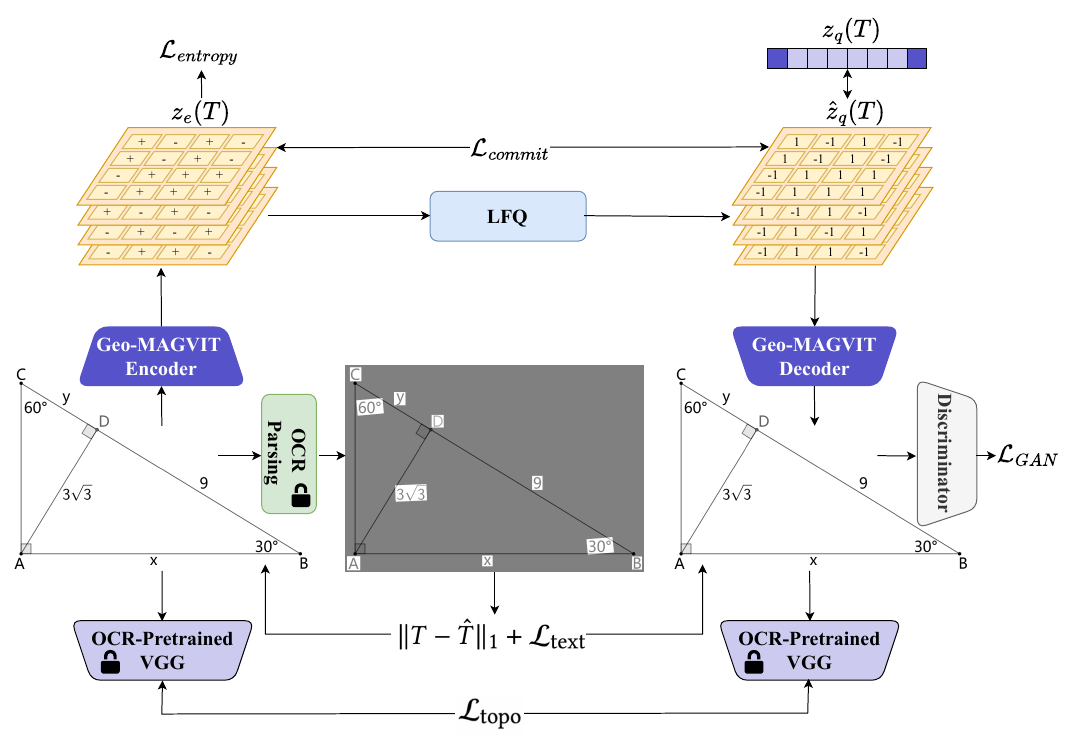}  % 调整图像宽度为 70% 的页面宽度
  \caption{Overview of Geo-MAGVIT}
  \Description{Geo-MAGVIT}
\end{figure}

Given a diagram \( T \in \mathbb{R}^{H \times W \times 3} \), we extract the representation after the Geo-MAGVIT Encoder, denoted as \( z_e(T) \in \mathbb{R}^{ H' \times W' \times \log(C)} \). We flatten it along the spatial dimension as \(z_{e}(T) = \{z_{e}^{i}(T) \}_{i=1}^{H'W'} \). Given a feature vector \( z_{e}^{i}(T) \in \mathbb{R}^{\log(C)} \), we apply Lookup-Free Quantization (LFQ), where the codebook becomes an integer set \( \mathbb{C} = \prod_{j=1}^{\log(C)} \{ -1, 1 \} \), and the latent space of each vector is decomposed as the Cartesian product. LFQ quantizes it according to the following equation:

\begin{equation}
\hat{z}_{q}^{i}(T) = \text{sign}(z_e^{i}(T)) = - \left[ z_e^{i}(T) \leq 0 \right] +   \left[ z_e^{i}(T) > 0 \right].
\end{equation}

This gives us \( \hat{z}_q(T) = \{\hat{z}_{q}^{i}(T)\}_{i=1}^{H'\times W'} \), and the reconstructed diagram is then obtained as:
\begin{equation}
    \hat{T} =  \mathcal{D}(z_{e}(T) + \text{sg}[\hat{z}_{q}(T) - z_{e}(T)]).
\end{equation}

Additionally, we can obtain the specific image token index representation $z_{q}(T)$ from $\hat{z}_{q}(T)$ as follows:
\begin{equation}
    z_{q}^{i}(T) = \sum_{j=1}^{\log(C)} 2^{j-2} \left[ \hat{z}_{q}^{i,j}(T) + 1 \right],\quad i=1,...,H'W'.
\end{equation}

The training incorporates multiple loss functions, including GAN loss, reconstruction loss, commit loss, and entropy loss, which are optimized as a weighted sum below:
\begin{equation}
\begin{aligned}
    \mathcal{L}_{\text{Geo-MAGVIT}} =  &\mathcal{L}_{\text{GAN}}
+ \lambda_{\text{rec}} \cdot \mathcal{L}_{rec} + \lambda_{\text{commit}} \cdot \mathcal{L}_{\text{commit}} \\
&+ \lambda_{\text{entropy}} \cdot \mathcal{L}_{\text{entropy}}. 
\end{aligned}
\end{equation}

We observe that MAGVIT encounters difficulties when reconstructing letters, numeric symbols and topological structures in the diagrams. To address this issue, we redesign the reconstruction loss \( \mathcal{L}_{\text{rec}} \) by incorporating both \( \mathcal{L}_{\text{topo}} \) and \( \mathcal{L}_{\text{text}} \) : 
\begin{equation}
    \mathcal{L}_{rec} = \| T - \hat{T} \|_1  + \mathcal{L}_{\text{topo}}+  \mathcal{L}_{\text{text}}.
\end{equation}

The topo-perceptual loss is to enhance the precision of the geometric topological structure in the generated diagrams. For implementation, we use the loss between features pre-trained on the VGG model for document OCR tasks \cite{rodriguez2023}, and it is formulated as follows:

%Inspired by OCR-VQGAN \cite{rodriguez2023}, We introduce the topo-perceptual loss to enhance the precision of the geometric topological structure in the generated diagrams. For implemnts, 我们使用针对文档ocr任务预训练的vgg的特征之间的损失来构建\cite{rodriguez2023}的 defined as:
\begin{equation}
\mathcal{L}_{\text{topo}} = \sum_{i=1}^{M} \| F_{\text{vgg}}^{(i)}(T) - F_{\text{vgg}}^{(i)}(\hat{T}) \|_1.
\end{equation}

We also introduce \( \mathcal{L}_{\text{text}} \) to improve the accuracy of textual reconstruction. We apply the OCR tool \cite{PaddleOCR} to generate bounding boxes for critical regions, such as endpoint labels and length/angle annotations on line segments within the diagrams. The design of \( \mathcal{L}_{\text{text}} \) is as follows:
\begin{equation}
\mathcal{L}_{\text{text}} = \left\| M \odot (T - \hat{T}) \right\|_1 .
\end{equation}

Besides, the original MAGVIT entropy loss is a convex function, and therefore always non-positive. We can prove that the minimum value of the entropy loss is $-\log(C)$ (proof in Appendix), hence we add a \( \log(C) \) term to guarantee  training stability:

%At the same time, the entropy loss can be shown to achieve a minimum value. To guarantee training stability, we add a \( \log(C) \) term:
%as proven in the appendix. To guarantee training stability, we add a \( \log(C) \) term:
\begin{equation}
   \mathcal{L}_{\text{entropy}} = \mathbb{E}\left[H[f(z_e(T))]\right] - H\left[\mathbb{E}[f(z_e(T))]\right] + \log(C).
\end{equation}
%And the commitment loss is shown as:
%\begin{equation}
%L_{\text{commit}} = \left\|\operatorname{sg}[z_q(T)] - z_e(T) \right\|
%\end{equation}

% f(z_e(T))是将z_e(T) (也可以看作logits), 转换成在codebook中的分布的算子.

\subsection{Multi-Task Instruction Tuning}
We initialize GeoUni using the weights of a pre-trained LLM and treat the multi-task instruction tuning as next-token prediction.

\subsubsection{Diagram Formalization Unified Prompting}
\begin{figure}[htbp]
    \centering
    \includegraphics[width=1.0\linewidth]{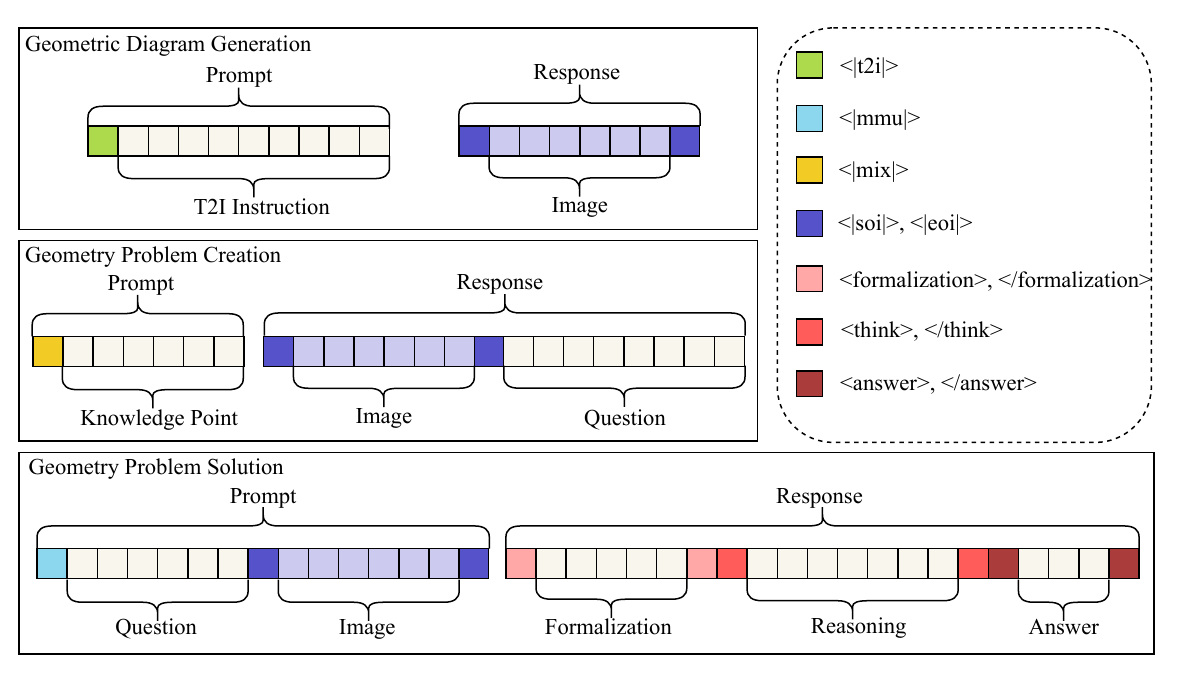}
    \caption{Diagram Formalization Unified Prompting}
    \Description{A figure showing the Geometry Unified Prompting.}
    \label{fig:geouni_prompting}
\end{figure}

To perform multi-task instruction tuning, we design the Diagram Formalization Unified Prompting to organize various types of data into a structured format. We pre-define three special tokens: \texttt{<|t2i|>}, \texttt{<|mmu|>}, and \texttt{<|mixing|>}, which represent the three tasks: text-to-diagram, problem-solving, and problem-generation. Additionally, \texttt{<|soi|>} and \texttt{<|eoi|>} are special tokens used to mark the start and end of discrete diagram tokens. \texttt{<|formalization|>} and \texttt{<|/formalization|>} are used to mark the beginning and end of the formalized description of the diagram. \texttt{<|think|>} and \texttt{<|/think|>} denote the start and end of the reasoning process in solving geometry problems, while \texttt{<|answer|>} and \texttt{<|/answer|>} mark the final answer. As shown in Figure~\ref{fig:geouni_prompting}, by adding different task tokens as the start of the sequence to distinguish different tasks, all data is converted into a 1D sequence of tokens.

\subsubsection{Training Objectives}
After processing with Geo-MAGVIT, we obtain the image tokens \( \mathbf{d} = \{ d_1, d_2, \dots, d_N \} \). The instruction and response text are also tokenized as \( \mathbf{t} = \{ t_1, t_2, \dots, t_M \} \) and \( \mathbf{r} = \{ r_1, r_2, \dots, r_K \} \), respectively. To perform unified next token prediction training, we employ three training objectives for the three different tasks. 

For the text-to-diagram task, we minimize the negative log-likelihood of the diagram tokens based on the instructions: 
\begin{equation}
    \mathcal{L}_{T2D} = \mathbb{E}_{( \mathbf{d},\mathbf{t})\sim D_{T2D}}[-\sum_{i=1}^{N} \log p_{\theta} (d_i| \mathbf{t}, d_1, \dots, d_{i-1})].
\end{equation}

For the geometry problem solution task, this task is a standard multi-modal understanding task, where the response answer is generated based on the provided problem text and diagram:
\begin{equation}
\mathcal{L}_{MMU} = \mathbb{E}_{( \mathbf{d},\mathbf{t},\mathbf{r})\sim D_{MMU}}[-\sum_{i=1}^{K} \log p_{\theta} (r_i \mid \mathbf{t}, \mathbf{d}, r_1, \dots, r_{i-1})] .
\end{equation}

For the geometry problem generation task, the process first generates the diagram and then the text, which involves mixing both text and image tokens. The loss function is defined as:
\begin{equation}
\begin{aligned}
    \mathcal{L}_{MIX} = & \mathbb{E}_{( \mathbf{d},\mathbf{t},\mathbf{r})\sim D_{MIX}}[-\sum_{i=1}^{N} \log p_{\theta} (d_i \mid \mathbf{t}, d_1, d_2, \dots, d_{i-1}) \\
                    & - \sum_{j=1}^{K} \log p_{\theta} (r_j \mid \mathbf{t}, \mathbf{d}, r_1, \dots, r_{j-1})].
\end{aligned}
\end{equation}

For joint training of the three tasks within a single framework, the total loss is defined as a weighted loss: 
\begin{equation}
\mathcal{L}_{GeoUni}=\lambda_{T2D}\cdot \mathcal{L}_{T2I} + \lambda_{MMU}\cdot \mathcal{L}_{MMU} + \lambda_{MIX}\cdot \mathcal{L}_{MIX} .
\end{equation}

\subsection{Reasoning Enhancement}
% During the reasoning enhancement phase, we employ reinforcement learning to further boost the model’s reasoning ability.
 Modality-specific functionality requires fine-tuning the base language model, which can hinder its original capabilities \cite{phi4mini2025}. Different tasks, like different modalities, present unique challenges. To address this, we employ GRPO and LoRA to fine-tune the reasoning adapter, enhancing reasoning performance without compromising the diagram generation capability of the instruction fine-tuned model. We redesign the reward function to better facilitate geometric reasoning tasks. The total reward function is defined as the sum of three components: format reward, formalization reward, and accuracy reward.

\textbf{Format Reward}
We encourage the model to structure its responses using a pre-defined format: \texttt{<formalization>} for diagram formalization, \texttt{<think>} for the reasoning process, and \texttt{<answer>} for the final answer. This reward is given a score of 1.0 if the response follows the above structure.

\textbf{Formalization Reward}  
Formalizing the diagram before reasoning helps the model better understand its geometric relationship. We supervise this process using a formalization score based on the Levenshtein distance between the predicted and ground truth \texttt{consCDL} and \texttt{imgCDL}, denoted as \( d_{\text{consCDL}} \) and \( d_{\text{imgCDL}} \), respectively. We define the individual scores as follows: 

\begin{align}
S_{\text{consCDL}} &= 1 - \frac{d_{\text{consCDL}}}{\max(|y_{\text{consCDL}}^*|, 1)}, \\
S_{\text{imgCDL}} &= 1 - \frac{d_{\text{imgCDL}}}{\max(|y_{\text{imgCDL}}^*|, 1)}.
\end{align}

The formalization reward is computed as the average of these two scores:

\begin{equation}
R_{\text{formal}}(y, y^*) = \max \left( 0, \frac{S_{\text{consCDL}} + S_{\text{imgCDL}}}{2} \right).
\end{equation}

\textbf{Accuracy Reward}
After performing formalization, which consists of structuring \texttt{consCDL} and \texttt{imgCDL}, the model proceeds to generate the reasoning process in natural language and the final answer. For both four-option multiple choice and open-ended questions, accuracy is measured based on whether the model's output exactly matches the standard answer.  

\section{Experiments}

\subsection{Datasets}
We train GeoUni on Formalgeo7K \cite{formalgeo} and SynthGeo228K \cite{zhang2025}. Each Formalgeo7K sample includes a bilingual description, a diagram, \texttt{consCDL} capturing topological relations, \texttt{imgCDL} for other geometric constraints, and a \texttt{formalSSS} symbolic solution translated into natural language. Considering the unique characteristics of geometry, we further design task-specific data augmentation strategies tailored for each task. For more details, please refer to the Appendix.

\subsection{Implementation Details}

The Geo-MAGVIT Encoder downsamples the input image resolution from 512 × 512 to 256 tokens and is trained for 50 epochs with a batch size of 16. For the base LLM model, we adopt DeepSeek-R1-Distill-Qwen-1.5B \cite{deepseekr1}. Multi-task instruction tuning is conducted for 50K steps with a batch size of 16. The Geo-Reasoning-Adapter is trained using LoRA with a rank of 256, applied to the q, k, v, and o projection modules. Training is performed with a batch size of 4 and a gradient accumulation step of 4. GRPO samples 8 responses per question and is trained for 4 epochs. All models are trained on 4 NVIDIA A800 (80GB) GPUs.

\subsection{Diagram Reconstruction}

\subsubsection{Metrics}
To evaluate the quality of diagram reconstruction, we design two types of metrics. One evaluates semantic accuracy in formal language, while the other focuses on pixel-level accuracy. 

\textbf{Geometry Semantic Matching Scores (GSMSs)} We use the geometric parser from~\cite{zhu2024fgeoparser} to translate the generated diagrams into these two CDL formats. Two precision metrics are proposed for evaluation: \textbf{Average Accuracy (AA)}, representing the average percentage of matched statements after transformation, and \textbf{Perfect Accuracy (PA)}, indicating the proportion of completely correct statements.
We compute AA and PA separately for \texttt{consCDL} (C-AA, C-PA) and \texttt{imgDDL} (I-AA, I-PA), as well as a combined Perfect Accuracy (CI-PA) that counts diagrams perfectly matching both CDLs.

\textbf{Geometry Pixel Matching Score (GPMS)}
Geometric diagrams are characteristically monochromatic and highly structured; only the black pixel regions encode geometric meaning, while the white background carries no task-relevant information. We define the geometric pixel sets as $F_{Gt}$ and $F_{Rec}$ where the pixels are black in the reference and reconstructed diagrams, respectively. The GPMS is then computed as: 
\begin{equation}
\text{GPMS} = 2 \times \frac{|F_{Gt} \cap F_{Rec}|}{|F_{Gt}| + |F_{Rec}|}.
\end{equation}

\subsubsection{Results}
Table~\ref{tab:reconstrution} provides a comparative analysis of diagram reconstruction performance across various image tokenizers, including MAGVIT \cite{luo2025openmagvit2} and our proposed Geo-MAGVIT. Geo-MAGVIT achieves superior results in both GSMSs and GPMS, demonstrating its strong ability to reconstruct diagrams with high fidelity, which is essential for the text-to-diagram task. While GPMS evaluates the accuracy at the pixel level, GSMSs focus on preserving the semantics of geometry. UniTok \cite{ma2025unitok} and QLIP \cite{zhao2025qlip} perform reasonably well on GSMSs, indicating that they can preserve the basic shapes of geometric diagrams. However, they fail to achieve accurate one-to-one reconstructions, often missing fine-grained details.

\begin{table}[htbp]
  \setlength{\tabcolsep}{2.5pt} 
  \renewcommand{\arraystretch}{0.8}
  \caption{Geometric Diagram Reconstruction Performance Comparison of Various Models Across Different Matrices}
  \label{tab:reconstrution}
  \begin{tabular}{l|ccccc|c}
    \toprule
    Model & C-AA & C-PA & I-AA & I-PA & CI-PA & GPMS \\
    \midrule
    UniTok\cite{ma2025unitok} & 81.43 & 54.29 & 69.95 & 48.94 & 31.44  & 39.65\\
    QLIP\cite{zhao2025qlip} & 77.82 & 50.63 & 68.73 & 48.10 & 30.06  & 52.87 \\
    MAGVIT\cite{luo2025openmagvit2} & 80.78 & 52.10 & 66.4 & 43.52 & 27.52 & 86.73 \\
    \midrule
    Geo-MAGVIT(Ours) & \textbf{83.10} & \textbf{55.05} & \textbf{75.98} & \textbf{55.71} & \textbf{35.24} & \textbf{91.32}\\
    \bottomrule
  \end{tabular}
\end{table}

\subsection{Text-To-Diagram}\label{sec:t2i_expr}

\subsubsection{Metrics}

In the text-to-diagram generation task, evaluating visual outputs is challenging due to the absence of pixel-level ground truth, making traditional image similarity metrics inapplicable. To address this, we adopt a symbolic evaluation approach. Specifically, we parse the generated diagrams into \texttt{consCDL} and \texttt{imgCDL} formats using a geometry parser as before, and compute BLEU-4 scores against the reference CDLs representations. This metric reflects structural fidelity without relying on pixel-level alignment and is effective in detecting inconsistencies in predicate composition, object relationships, and the symbolic structure of the diagrams.

\begin{table*}[t]
  \caption{Text-To-Diagram Performance Comparison of Various Models Across Different Matrices}
  \renewcommand{\arraystretch}{0.7}
  \label{tab:T2D}
  \begin{tabular}{c|ccc|ccc|ccc|ccc}
    \toprule
    \multirow{2}{*}{Model} & \multicolumn{3}{c|}{Construction/EN} & \multicolumn{3}{c|}{Image/EN} & \multicolumn{3}{c|}{Construction/CN} & \multicolumn{3}{c}{Image/CN} \\
    \cmidrule(lr){2-4} \cmidrule(lr){5-7} \cmidrule(lr){8-10} \cmidrule(lr){11-13}
    & Caption & CDL & GPT & Caption & CDL & GPT & Caption & CDL & GPT & Caption & CDL & GPT \\
    \midrule
    Show-o\cite{xie2024showo} & 29.39 & 25.01 & 24.84 & 34.51 & 24.58 & 34.01 & 32.67 & 29.82 & 35.63 & 33.80 & 31.33 & 34.00 \\
    Janus-Pro-7B\cite{chen2025januspro} & 14.55 & 17.72 & 20.85 & 30.22 & 29.7 & 33.02 & 34.85 & 35.47 & 34.69 & 29.09 & 20.56 & 26.55 \\
    Anole-7B\cite{chern2024anole} & 16.84 & 18.68 & 13.63 & 24.11 & 30.73 & 23.73 & 12.42 & 19.55 & 14.25 & 34.01 & 28.59 & 34.40 \\
    Emu3\cite{wang2024emu3} & 22.06 & 18.76 & 16.00 & 20.89 & 19.88 & 19.90 & 21.20 & 25.25 & 18.78 & 20.96 & 22.65 & 19.75 \\
    Unified-IO\cite{lu2022unifiedio} & 32.27 & 29.37 & 31.63 & 17.84 & 16.41 & 16.74 & 37.62 & 33.81 & 32.51 & 27.95 & 19.43 & 29.44 \\
    SEED-X\cite{seedx} & 34.20 & 31.78 & 27.84 & 14.80 & 10.47 & 12.30 & 30.78 & 31.40 & 27.40 & 15.60 & 11.37 & 14.89 \\
    \midrule
    PixArt-$\Sigma$\cite{chen2024pixartsigma} & 18.51 & 17.94 & 17.14 & 19.35 & 19.44 & 15.39 & 28.55 & 23.67 & 28.29 & 17.89 & 21.40 & 16.45 \\
    SD-v1.5\cite{Rombach2022}& 21.69 & 19.51 & 23.06 & 22.46 & 20.71 & 22.26 & 25.03 & 25.14 & 27.30 & 14.11 & 13.34 & 16.72 \\
    SDXL-Turbo\cite{sdxl} & 12.94 & 11.12 & 14.76 & 13.86 & 15.93 & 11.76 & 30.36 & 17.86 & 29.62 & 19.52 & 10.84 & 18.38 \\
     DALL-E-2\cite{dalle2} & 24.38 & 19.75 & 18.30 & 28.29 & 24.74 & 19.27 & 22.50 & 19.56 & 20.78 & 20.31 & 21.36 & 17.47 \\
     \midrule
    GeoUni-1.5B(Ours) & \textbf{73.00} & \textbf{73.43} & \textbf{72.41} & \textbf{78.46} & \textbf{79.65} & \textbf{77.53} & \textbf{73.72} & \textbf{73.00} & \textbf{72.77} & \textbf{79.40} & \textbf{79.54} & \textbf{77.82} \\
    \bottomrule
  \end{tabular}
\end{table*}

\begin{figure}[htbp]
    \centering
    \includegraphics[width=1.0\linewidth]{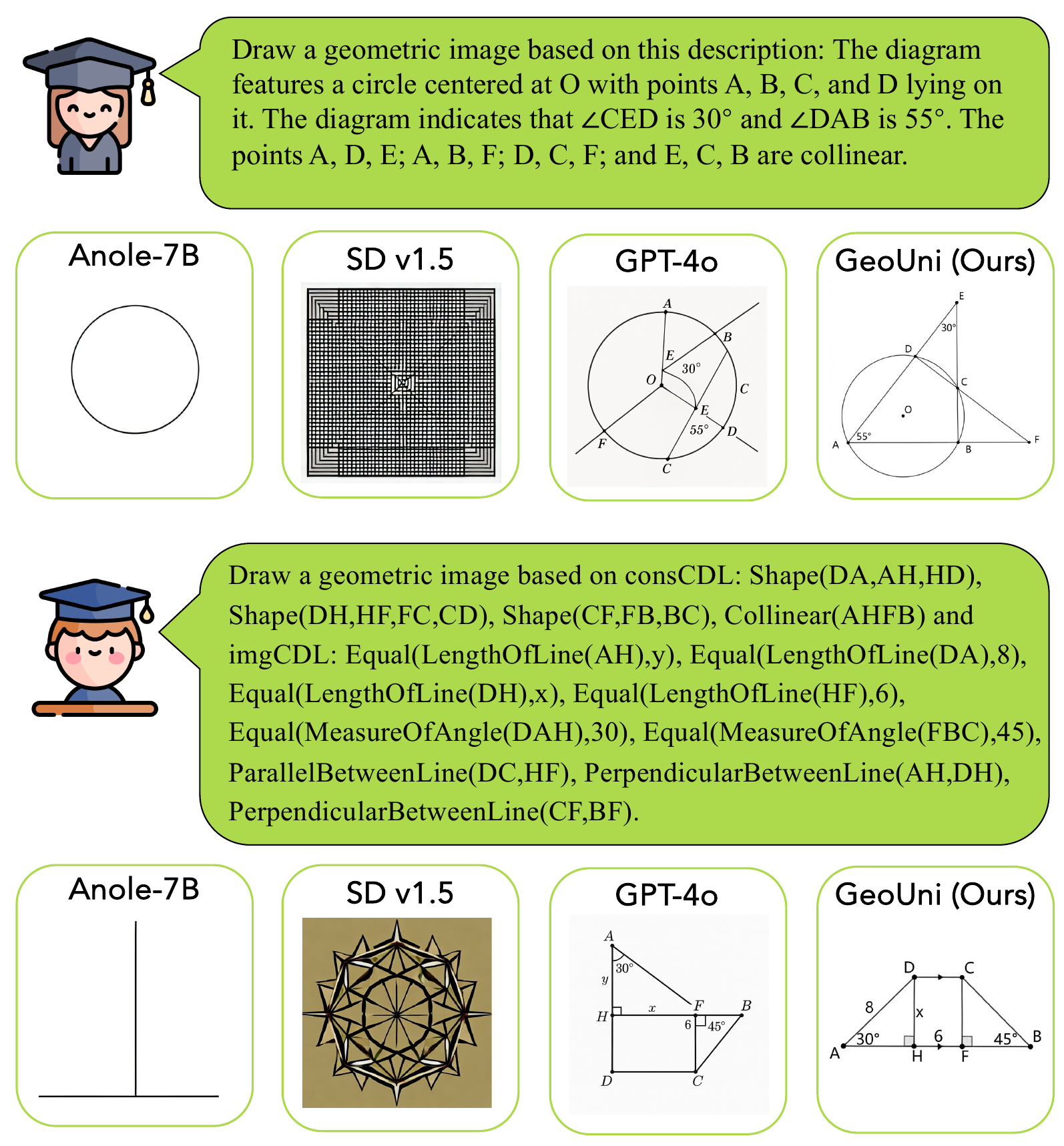}
    \caption{Text-to-Diagram}
    \label{fig:t2i}
    \Description{A Showcase of Text to Diagram.}
\end{figure}

\subsubsection{Results}

In Table~\ref{tab:T2D}, we show a comparative analysis of text-to-diagram generation performance across unified models, text-to-image (T2I) models, and our proposed GeoUni, under three types of prompts: natural language captions, formalized CDL descriptions, and GPT-rewritten instructions, in both English and Chinese. GeoUni consistently achieves the highest scores across all settings, significantly outperforming baselines in both \texttt{consCDL} and \texttt{imgCDL} BLEU-4 metrics. It is important to note that BLEU-4 is a relatively soft metric. Although some unified and T2I models obtain moderate scores, their generated outputs often fail to resemble valid geometric diagrams in structure or semantics.

As shown in Figure~\ref{fig:t2i}, SD v1.5 \cite{Rombach2022} tends to generate visually rich or stylized content, but its outputs lack geometric structure. Anole-7B \cite{chern2024anole}, fine-tuned from Chameleon-7B \cite{chameleon}, is capable of generating simple closed shapes such as circles, but lacks the ability to execute more complex geometry instructions. Additionally, although the recently released GPT-4o \cite{gpt4o} demonstrates impressive image generation capabilities, it currently lacks API support. Preliminary results using the web interface show that it can generate visually clear geometric figures, but the outputs do not satisfy precise geometric constraints. 

\subsection{Reasoning}
\subsubsection{Metrics}
To comprehensively evaluate the geometric reasoning capabilities of the models, we test their performance on both multiple choice and open-ended questions using three public datasets: Formalgeo7K, Geometry3K \cite{lu2021inter}, and GeoQA \cite{geoqa}. For both question types, the models are instructed to reason step-by-step, generate detailed solutions, and present answers in a standardized format to allow accurate comparison with reference solutions. Accuracy metrics are calculated separately based on the question type (multiple choice vs. open-ended) and language (English vs. Chinese), represented as EN-C, CN-C, EN-OE, and CN-OE, respectively.

\begin{table*}[t]
 \setlength{\tabcolsep}{4.0pt} 
 \renewcommand{\arraystretch}{0.7}
  \caption{Geometric Reasoning Performance Comparison of Various Models Across Different Matrices}
  \label{tab:reasoning}
  \begin{tabular}{l|cccc|cccc|cccc}
    \toprule
    \multirow{2}{*}{Model} & \multicolumn{4}{c|}{Formalgeo7K} & \multicolumn{4}{c|}{Geometry3k} & \multicolumn{4}{c}{GeoQA}\\
       & EN-C & EN-OE & CN-C & CN-OE & EN-C & EN-OE & CN-C & CN-OE & EN-C & EN-OE & CN-C & CN-OE  \\
    \midrule
    Show-o\cite{xie2024showo} & 17.90 & 2.57 & 16.57 & 0.29 & 16.44 & 2.31 & 17.59 & 0.46 & 18.93 & 2.75 & 15.86 & 0.16 \\
    Chameleon-7B\cite{chameleon} & 10.38 & 5.90 & 11.81 & 3.62 & 9.03 & 3.94 & 11.34 & 3.01 & 11.33 & 7.28 & 12.14 & 3.01 \\
    Janus-Pro-7B\cite{chen2025januspro} & 25.62 & 13.05 & 8.38 & 10.57 & 23.38 & 12.04 & 6.25 & 10.65 & 27.18 & 13.75 & 9.87 & 10.52 \\
    Emu3\cite{wang2024emu3} & 45.24 & 7.14 & 21.33 & 3.52 & 40.05 & 3.94 & 21.53 & 2.08 & 48.87 & 9.39 & 21.20 & 4.53 \\
    SEED-X\cite{seedx} & 13.90 & 1.33 & 7.24 & 3.43 & 11.81 & 0.69 & 5.09 & 1.85 & 15.37 & 1.78 & 8.74 & 4.53 \\
    GPT-4o\cite{gpt4o}  & 47.52 & 20.10 & 49.71 & 28.48 & 50.69 & 23.38 & 52.08 & 28.70 & 45.31 & 17.80 & 48.06 & 28.32\\
    \midrule
    G-LLaVA-13B\cite{gao2023gllava} & 47.81 & 14.10 & - & - & 5.56 & 34.95 & - & -  & 56.80 & 20.06 & - & -\\
    Qwen2.5-VL-7B\cite{qwen25vl} & 61.71 & 26.38 & 60.48 & 27.14 & 64.12 & 23.84 & 59.26 & 22.45 & 60.03 & 28.16 & 61.33 & 30.42\\
    QWQ-72B\cite{qvq-72b-preview} & 51.62 & 16.57 & 20.19 & 7.14 & 50.69 & 23.61 & 18.52 & 7.41 & 52.27 & 11.65 & 21.36 & 6.96 \\
    Phi-4-Multimodal\cite{phi4mini2025} & 48.19 & 14.48 & 43.24 & 10.00 & 52.78 & 10.88 & 39.81 & 9.03 & 44.98 & 16.99 & 45.63 & 10.68\\
    \midrule
    Qwen2.5-Math-1.5B\cite{qwen25math} & 48.95 & 24.12 & 47.43 & 20.86 & 45.60 & 20.14 & 47.45 & 21.99 & 51.29 & 26.90 & 47.41 & 20.06 \\
    DS-R1-Distill-Qwen-1.5B\cite{deepseekr1} & 40.38 & 16.57 & 38.48 & 11.71 & 37.50 & 19.68 & 38.19 & 16.20 & 42.39 & 14.40 & 38.67 & 8.58\\
    Qwen2.5-Math-72B\cite{qwen25math} & 70.67 & 37.90 & 67.90 & 37.62 & 69.44 & 36.81 & 65.05 & 33.33 & 71.52 & 38.67 & 67.90 & 40.61 \\
    DeepSeek-R1\cite{deepseekr1} & 64.86 & 31.05 & \textbf{77.24} & 31.71 & 66.90 & 36.34 & \textbf{75.23} & 36.57 & 63.43 & 27.35 & \textbf{78.64} & 28.32\\
    DeepSeek-V3\cite{deepseekv3} & 70.29 & 33.46 & 65.71 & 26.38 & 71.30 & 35.66 & 63.19 & 28.01 & 69.58 & 30.32 & 67.48 & 25.24\\
    \midrule
    GeoUni-1.5B(Ours) & \textbf{75.43} & \textbf{59.81} & 73.52 & \textbf{55.33} & \textbf{71.76} & \textbf{50.00} & 69.68 & \textbf{45.27} & \textbf{77.99} & \textbf{66.67} & 76.21 & \textbf{62.30} \\
    \bottomrule
  \end{tabular}
\end{table*}

\subsubsection{Results} 

Table~\ref{tab:reasoning} summarizes the reasoning performance across three benchmark datasets in unified models, MLLMs, LLM, and our proposed GeoUni with only a 1.5B-parameter LLM. GeoUni achieves the highest accuracy on English multiple choice questions—75.43\%, 71.76\%, and 77.99\% on Formalgeo7K, Geometry3K, and GeoQA respectively—outperforming significantly larger models such as DeepSeek-V3 \cite{deepseekv3} and DeepSeek-R1 \cite{deepseekr1}. While it slightly lags behind DeepSeek-R1 and Qwen2.5-VL-32B \cite{qwen25vl} in Chinese multiple choice settings, GeoUni remains competitive overall.

On open-ended tasks, GeoUni demonstrates clear advantages in both English and Chinese, particularly through its step-by-step reasoning presented in structured answer formats. For example, on Formalgeo7K (CN-OE), it reaches 55.33\%, far surpassing DeepSeek-R1’s 31.71\%.

We also observe that many unified models (e.g., Show-o \cite{xie2024showo}, Janus-Pro \cite{chen2025januspro}, and Emu3 \cite{wang2024emu3}) perform poorly in both multiple choice and open-ended formats. These models often fail to follow task instructions consistently, leading to performances even worse than random guessing in four-option multiple choice settings. Moreover, models like G-LLaVA-13B \cite{gao2023gllava} do not support Chinese input and thus are only evaluated on English subsets.

\begin{figure}[htbp]
    \centering
    \includegraphics[width=1.0\linewidth]{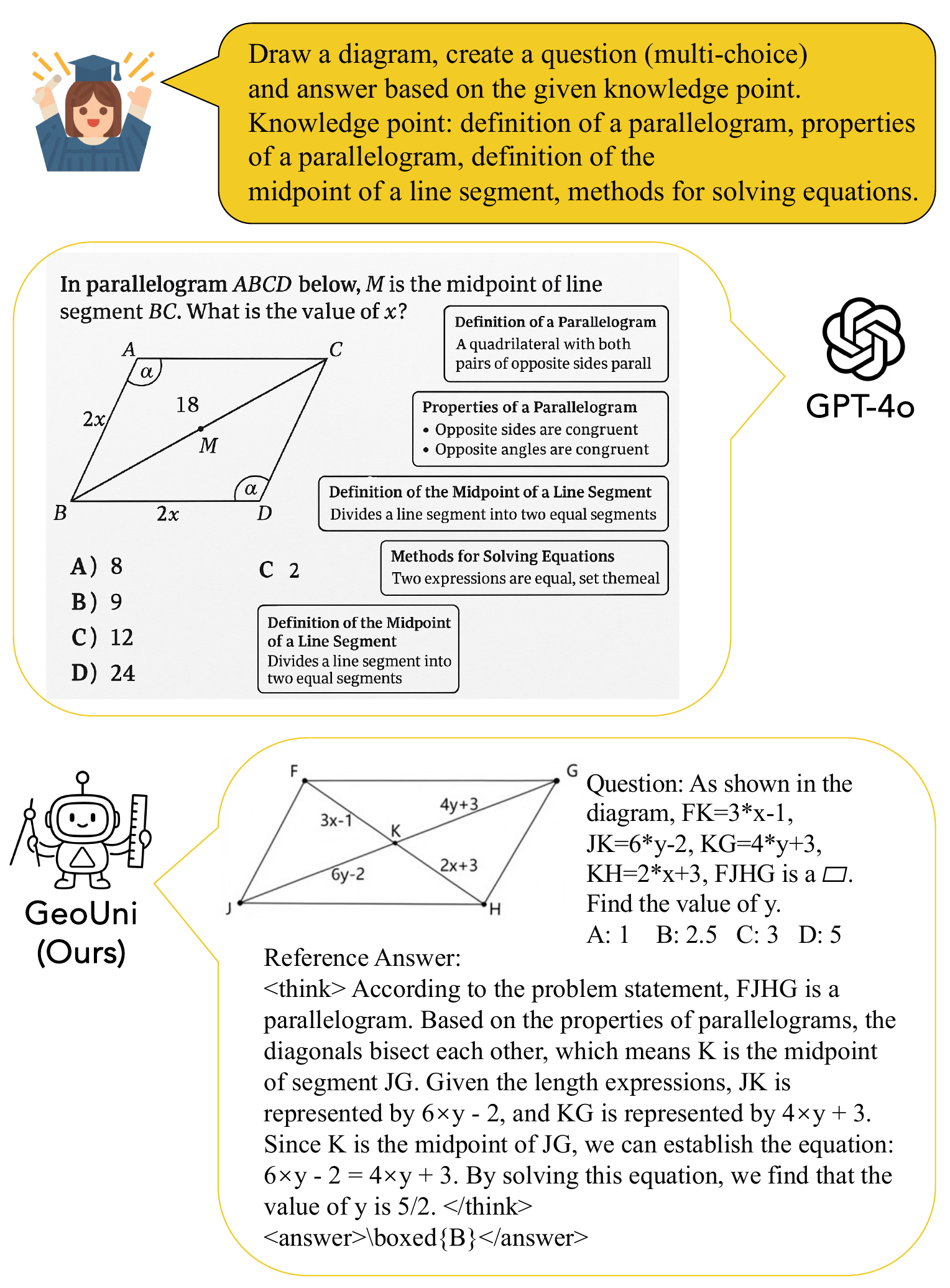}
    \caption{Problem Creation}
    \label{fig:problem}
    \Description{A Showcase of Problem Creation.}
\end{figure}

\subsection{Problem Creation}

We further examine GeoUni's problem-generation capability by prompting it with geometric knowledge points to generate corresponding problems along with appropriate diagrams. As shown in the comparative example between GPT-4o \cite{gpt4o} and GeoUni in Figure ~\ref{fig:problem}, although GPT-4o can generate images that align with the instructions, the resulting geometric problem is actually unsolvable. In contrast, GeoUni not only generates meaningful geometry problems but also produces accurate geometric diagrams and provides detailed reference answers.

\subsection{Ablation Studies}
\subsubsection{Geo-MAGVIT}

\begin{figure}[htbp]
    \centering
    \includegraphics[width=1.0\linewidth]{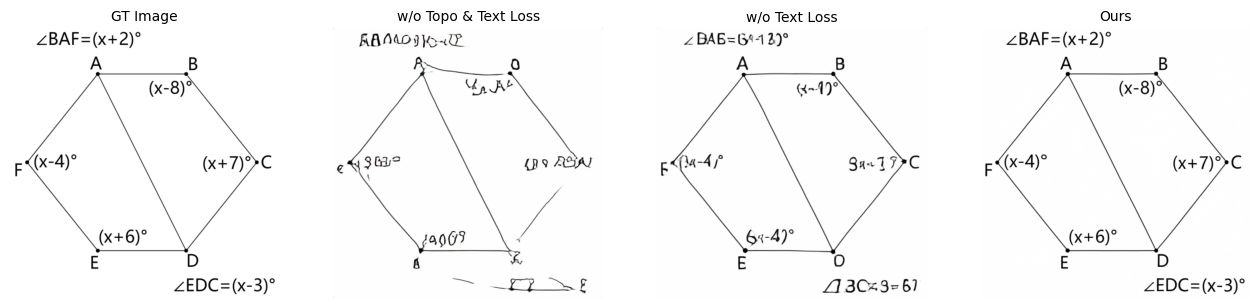}
    \caption{Ablation Study of Geo-MAGVIT}
    \label{fig:ablation-1}
    \Description{A Showcase of Loss Design of Geo-MAGVIT.}
\end{figure}

We investigate the impact of two key training objectives in Geo-MAGVIT: the topo-perceptual reconstruction loss ($\mathcal{L}_{\text{topo}}$) and the text reconstruction loss ($\mathcal{L}_{\text{text}}$), as shown in Table~\ref{tab:ablation-1}. We evaluate the model’s performance under different configurations by removing either or both losses during training. Removing $\mathcal{L}_{\text{text}}$ significantly impairs the reconstruction of textual elements in the diagram, such as endpoint labels and angle annotations. When both $\mathcal{L}_{\text{topo}}$ and $\mathcal{L}_{\text{text}}$ are removed, the model’s ability to preserve the overall geometric structure degrades notably, as illustrated in Fig.~\ref{fig:ablation-1}.

\begin{table}[htbp]
  \renewcommand{\arraystretch}{0.7}
  \caption{Impact of Different Training Loss on Geo-MAGVIT}
  \label{tab:ablation-1}
  \begin{tabular}{cc|cccccc}
    \toprule
    Topo & Text & C-AA & C-PA & I-AA & I-PA & CI-PA & GPMS \\
    \midrule
    - & - & 5.86 & 1.90 & 3.28 & 2.67 & 0.00 & 58.66 \\
    \checkmark & - & 43.24 & 26.38 & 18.78 & 8.00 & 4.48 & 71.11 \\   
    \checkmark & \checkmark & \textbf{83.10} & \textbf{55.05} & \textbf{75.98} & \textbf{55.71} & \textbf{35.24} & \textbf{91.32}  \\
    \bottomrule
  \end{tabular}
\end{table}

\subsubsection{Reasoning}
We examine the effects of the stage 3 reasoning enhancement and formalization of diagram information before solving the problem in Formalgeo7K. Table~\ref{tab:ablation-2} shows that removing the reasoning enhancement (GRPO) in a significant drop in performance across all settings, confirming its importance in boosting model reasoning ability.  Likewise, omitting the formalization step also leads to a decrease in both English and Chinese accuracy, especially in open-ended settings. 

\begin{table}[htbp]
    \renewcommand{\arraystretch}{0.7}
    \caption{Impact of GRPO and formalization on our model's performance on the Formalgeo7K}
    \centering
    \begin{tabular}{c|cccc}
        \toprule
        \multirow{2}{*}{Model} & \multicolumn{4}{c}{Accuracy} \\
          & EN-C & EN-OE & CN-C & CN-OE \\
        \midrule
        GeoUni (w/o GRPO) & 53.81 & 39.90 & 55.52 & 39.33 \\   
        GeoUni (w/o Formalization) & 70.57 & 53.33 & 70.19 & 52.86 \\ 
        GeoUni & \textbf{75.43} & \textbf{59.81} & \textbf{73.52} & \textbf{55.33} \\  
        \bottomrule
    \end{tabular}
    \vspace{-5pt}
    \label{tab:ablation-2}
\end{table}

\section{Conclusion}
In this paper, we propose a unified geometry expert model, \textbf{GeoUni}, which integrates geometry problem solving, diagram generation, and problem creation within a single framework. Extensive experimental results demonstrate that GeoUni outperforms existing models in all three tasks. Most importantly, GeoUni makes geometry problem creation a practical reality, bridging the gap between problem solving and teaching. In future works, we aim to explore acceleration techniques to further improve the efficiency of GeoUni, enabling faster geometry problem generation.

\bibliographystyle{ACM-Reference-Format}
\bibliography{refs}

\appendix

\section{Dataset Details}
\begin{table*}[htbp]
    \centering
    \setlength{\tabcolsep}{2pt} 
    \caption{Details of training datasets at different stages.}
    \begin{tabular}{c|c|c|c}
        \toprule
        \multicolumn{2}{c|}{Stage}  & Sources & Training Samples  \\
        \midrule
        \multicolumn{2}{c|}{Diagram Tokenization Pretraining} & Formalgeo7K (train set) + SynthGeo228K (sampled)  & $5,950 + 193,204 = 199,254$ \\
        \midrule
        \multirow{3}{*}{Multi-Task Instruction Tuning} & T2D & Formalgeo7K (train set) + SynthGeo228K (sampled) & $(4 \times 10 + 2) \times 5,950 + 130,900 = 380,800$ \\
         & MMU & Formalgeo7K (train set sampled) & $(4 \times 8) \times 4,900 = 156,800$ \\
         & MIX & Formalgeo7K (train set sampled) & $(4 \times 8) \times 4,900 = 156,800$ \\
         \midrule
         \multicolumn{2}{c|}{Reasoning Enhancement} & Formalgeo7K (train set sampled) & $1,050 \times 8 = 8,400$ \\
         \bottomrule
    \end{tabular}
    \label{tab:dataset details}
\end{table*}

\begin{figure}[htbp]
    \centering
    \includegraphics[width=1.0\linewidth]{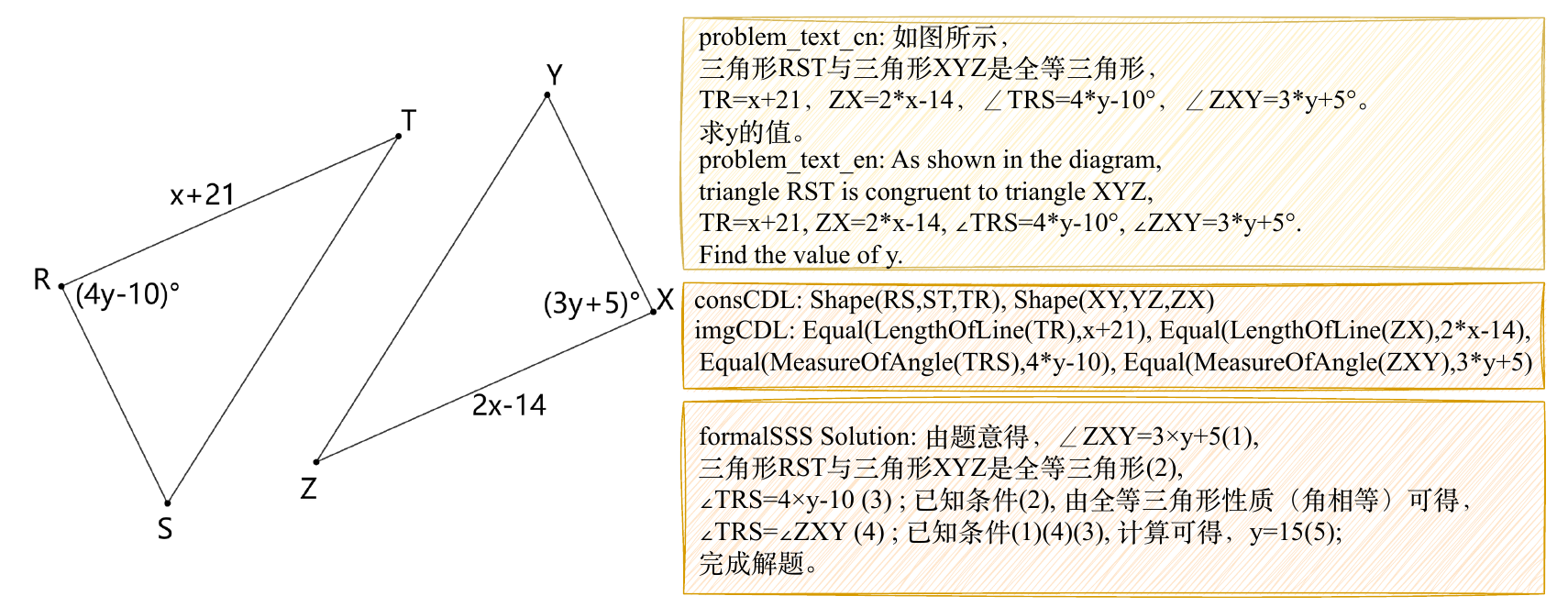}
    \caption{Data Sample of Formalgeo7K.} 
    \label{fig:data_sample_formalgeo}
    \Description{Showcase of formalgeo}
\end{figure}

\begin{figure}[htbp]
    \centering
    \includegraphics[width=1.0\linewidth]{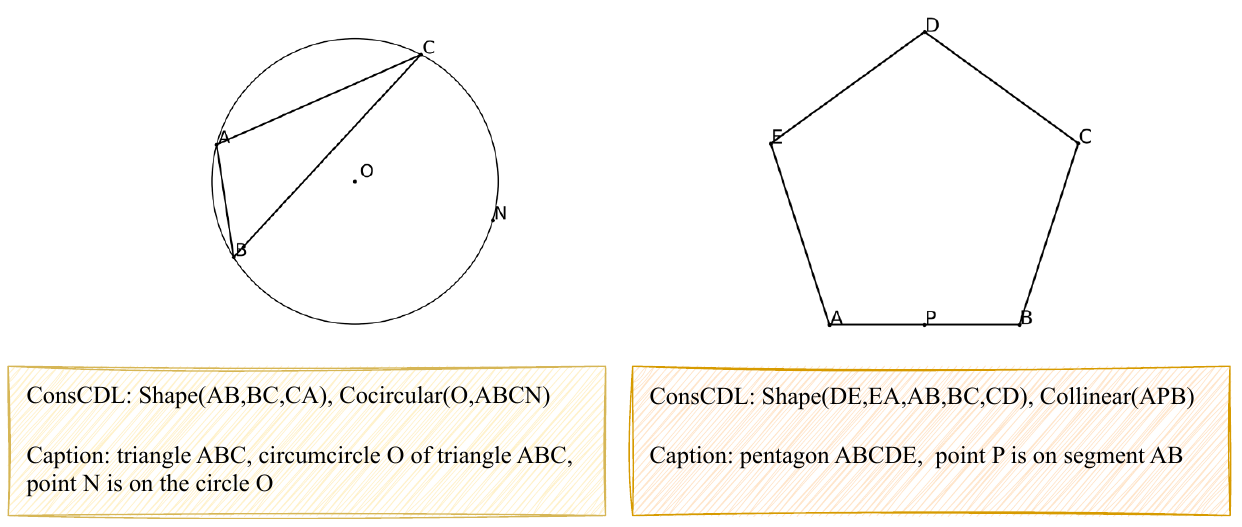}
    \caption{Data Sample of SynthGeo228K.}
    \label{fig:data_sample_syth}
    \Description{Showcase of SynthGeo228K}
\end{figure}

Our GeoUni model is trained on two datasets: FormalGeo7K \cite{formalgeo} and SynthGeo228K \cite{zhang2025}. The FormalGeo7K dataset contains 5,950 training samples and 1,050 test samples. From the training set, we further separate 1,050 samples specifically for training the Reasoning Enhancement module. SynthGeo228K is a large-scale synthetic dataset comprising geometric diagrams paired with corresponding descriptions. A sample from the FormalGeo7K dataset is shown in Figure~\ref{fig:data_sample_formalgeo}. Each instance includes \texttt{problem-text-cn} and \texttt{problem-text-en} as the Chinese and English problem descriptions, respectively; \texttt{consCDL} and \texttt{imgCDL} as structured diagram representations; and \texttt{formalSSS-Solution}, which presents a symbolic reasoning process expressed in natural language. Additionally, Figure~\ref{fig:data_sample_syth} displays an example from the SynthGeo228K dataset, where each image is accompanied by a formal \texttt{consCDL} representation and a natural language caption.

For diagram tokenization pretraining, we utilize diagrams from the Formalgeo7K training set, which includes 5,950 diagrams, and randomly sample 193,304 diagrams from SynthGeo228K.  

\begin{figure}[htbp]
    \centering
    \includegraphics[width=1.0\linewidth]{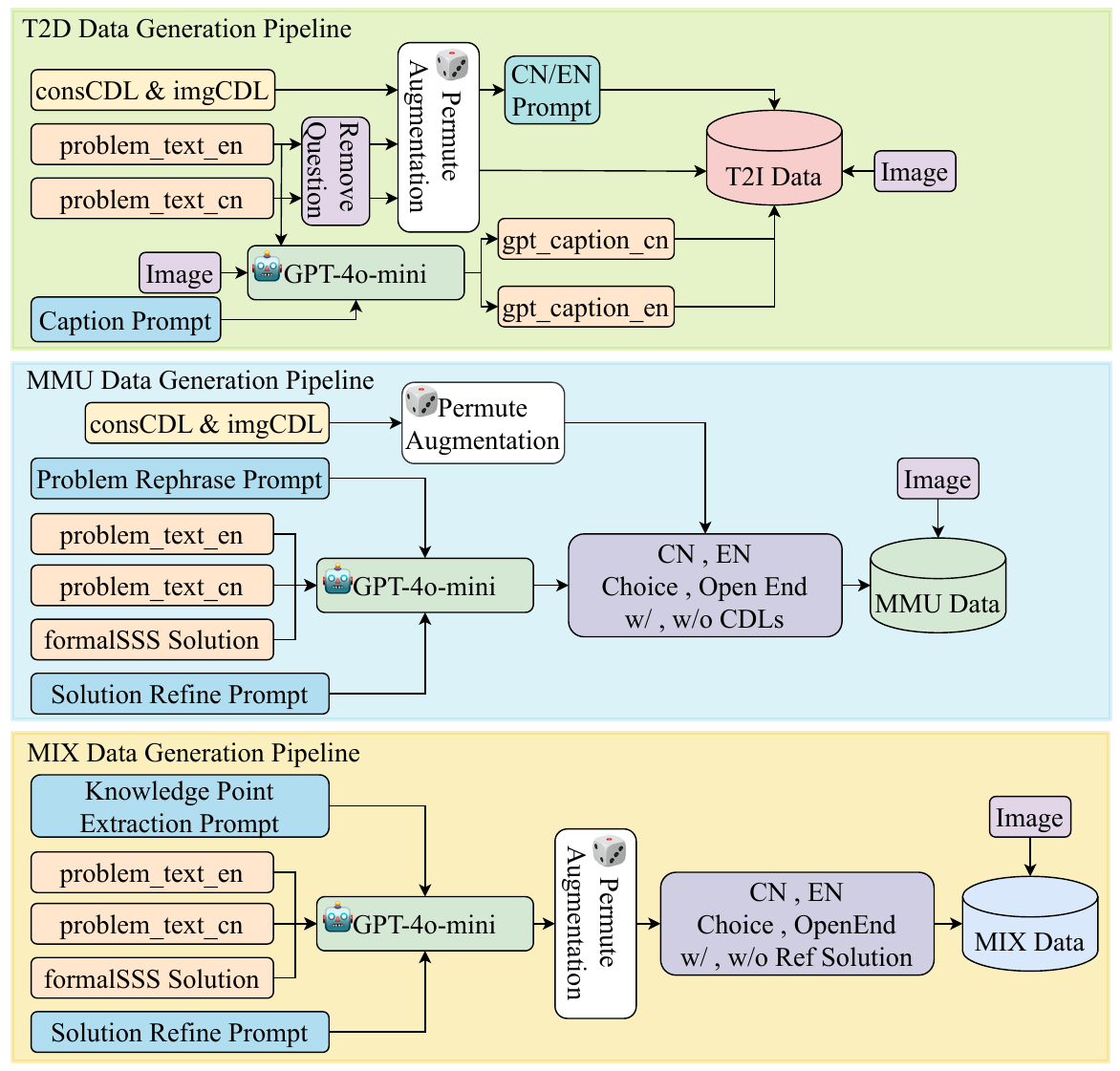}
    \caption{Data Augmentation Pipeline for Formalgeo7K.}
    \Description{A figure showing the Data Augmentation pipeline for Formalgeo7K.}
    \label{fig:data_generation}
\end{figure}

Figure~\ref{fig:data_generation} illustrates the data augmentation pipeline for multi-task instruction tuning based on the Formalgeo7K dataset. For the Text-to-Diagram (T2D) task, we exploit the permutation invariance of geometric descriptions in the CDLs and the problem text (with questions removed) to perform 10× permutation-based augmentation. We construct both Chinese and English prompt templates to guide diagram generation. Additionally, we employ GPT-4o-mini to generate diagram descriptions conditioned on both the image content and the associated problem text. This results in 249,900 augmented samples—calculated as (4 prompt templates × 10 permutations + 2 GPT-generated descriptions) × 5,950 samples. An additional 130,900 synthetic samples are incorporated, yielding a total of 380,800 samples for T2D training.

For the geometry problem solving (MMU) task, we define 8 distinct question-answering modes by considering language (Chinese or English), question type (multiple-choice or open-ended), and whether geometric diagram formalization is required prior to answering. To enhance the diversity of question phrasing, we apply a 4× rephrasing strategy to each question using GPT-4o-mini. For modes involving pre-formalized geometric diagrams, we further adopt sequence-level augmentation on the corresponding CDL representations. Given that the original \texttt{formalSSS-solutions} are not well-suited for direct model training, we employ GPT-4o-mini to refine them into more human-like solution processes. From the original training set of 5,950 samples, we resample 4,900 instances and construct a dataset of 156,800 MMU samples via the data augmentation pipeline, incorporating all 8 modes and 4 rephrasings per instance.

For the problem augmentation (MIX) task, which involves problem augmentation, we consider eight distinct modes based on language (Chinese or English), question type (multiple-choice or open-ended), and the presence or absence of a reference answer. We employ GPT-4o-mini to extract knowledge points from the original problem text and the corresponding \texttt{formalSSS} solutions. Given the order-invariant nature of these knowledge points, we apply a 4× permutation-based augmentation strategy. Furthermore, similar to the MMU dataset, all reference answers are refined using GPT-4o-mini. In total, our MIX dataset contains (4 × 8) × 4900 = 156,800 samples after augmentation. 

The reasoning enhancement dataset comprises the remaining 1,050 samples from the Formalgeo7K training set in the MMU dataset, which are used for MMU and MIX tasks. By applying various configurations—including question type (multiple-choice or open-ended), language (Chinese or English), and whether formalization is performed beforehand—a total of 8× augmented reinforcement learning samples were constructed, resulting in 8,400 RL training instances.

\section{Mathematical Proof}
The original entropy loss\cite{luo2025openmagvit2} defined as
\[
\mathcal{L}_{\text{entropy}}^{\text{old}} = \mathbb{E}\left[H(f(z_e(T)))\right] - H\left(\mathbb{E}[f(z_e(T))]\right)
\]
is always non-positive, i.e., \(\mathcal{L}_{\text{entropy}}^{\text{old}} \le 0\), and reaches its minimum value \(-\log(C)\), where \(C\) denotes the number of codebook entries, and \(f(\cdot)\) maps feature representations into a probability distribution over the codebook.

\begin{proof}
Let us denote 
$$\mathbf{p}(T) = f(z_e(T)) = (p_1(T),p_2(T), \dots, p_C(T)),$$
representing the probability distribution over the codebook for a given input \(T\). Then the entropy loss becomes:
\begin{equation}
   \mathcal{L}_{\text{entropy}}^{\text{old}} = \mathbb{E}_T\left[H(\mathbf{p}(T))\right] - H\left(\mathbb{E}_T[\mathbf{p}(T)]\right).
\end{equation}
Note that for \(p \in (0, 1)\), the function \(g(p) = -p \log (p)\) has a second derivative \(g''(p) = -\frac{1}{p} < 0\), implying that \(g(p)\) is convex. Therefore, the entropy function
\begin{align}
    H(\mathbf{p}) &= \sum_{i=1}^{C} -p_i \log(p_i) \\
                  &= \sum_{i=1}^{C} g(p_i)
\end{align}
is a sum of convex functions, and hence also convex. By Jensen's inequality, we obtain:
\begin{equation}
    \mathbb{E}_T\left[H(\mathbf{p}(T))\right] \le H\left(\mathbb{E}_T[\mathbf{p}(T)]\right),
\end{equation}
which leads to \(L_{\text{entropy}}^{\text{old}} \le 0\), indicating a contradiction with the common expectation that loss functions are non-negative.

Furthermore, using the property of entropy that \(0 \le H(\mathbf{p}) \le \log(C)\), we derive:
\begin{align}
    \mathbb{E}_T\left[H(\mathbf{p}(T))\right] &\ge 0, \\
    -H\left(\mathbb{E}_T[\mathbf{p}(T)]\right) &\ge -\log(C).
\end{align}
Summing the two inequalities yields:
\begin{equation}
    \mathbb{E}_T\left[H(\mathbf{p}(T))\right] - H\left(\mathbb{E}_T[\mathbf{p}(T)]\right) \ge -\log(C),
\end{equation}
i.e., \(\mathcal{L}_{\text{entropy}}^{\text{old}} \ge -\log(C)\).

To demonstrate that the lower bound is tight, we consider a specific case. Suppose we have B = C samples, and the sample mean is used to approximate the expectation. In this case, the loss becomes:
\begin{equation}
    \mathcal{L}_{\text{entropy}}^{\text{old},*} = \frac{1}{B} \sum_{k=1}^{B} H(\mathbf{p}(T_k)) - H\left(\frac{1}{B} \sum_{k=1}^{B} \mathbf{p}(T_k)\right).
\end{equation}
For $k=1,..,C$, let us define
\begin{equation}
    \mathbf{p}(T_k) = (0, \dots, 0, 1, 0, \dots, 0),
\end{equation}
where the 1 appears in the \(k\)-th position, and the rest are zeros. In this case, we have:
\begin{align}
    H(\mathbf{p}(T_k)) &= 0,\\
    H\left(\frac{1}{B} \sum_{k=1}^{B} \mathbf{p}(T_k)\right) &= H\left(\frac{1}{C}, \dots, \frac{1}{C}\right) \\
    &= \log(C).
\end{align}
Therefore,
\begin{align}
    \mathcal{L}_{\text{entropy}}^{\text{old},*} &= \frac{1}{B} \sum_{k=1}^{B} H(\mathbf{p}(T_k)) - H\left(\frac{1}{B} \sum_{k=1}^{B} \mathbf{p}(T_k)\right) \\
    &= 0 - \log(C) \\
    &= -\log(C).
\end{align}
This concludes the proof that the minimum of \(\mathcal{L}_{\text{entropy}}^{\text{old}}\) is \(-\log(C)\).
\end{proof}

\section{Examples for Diagram Reconstruction}
\begin{figure*}[htbp]  % 使用 figure* 来跨两栏显示图像
    \centering
    \includegraphics[width=0.9\linewidth]{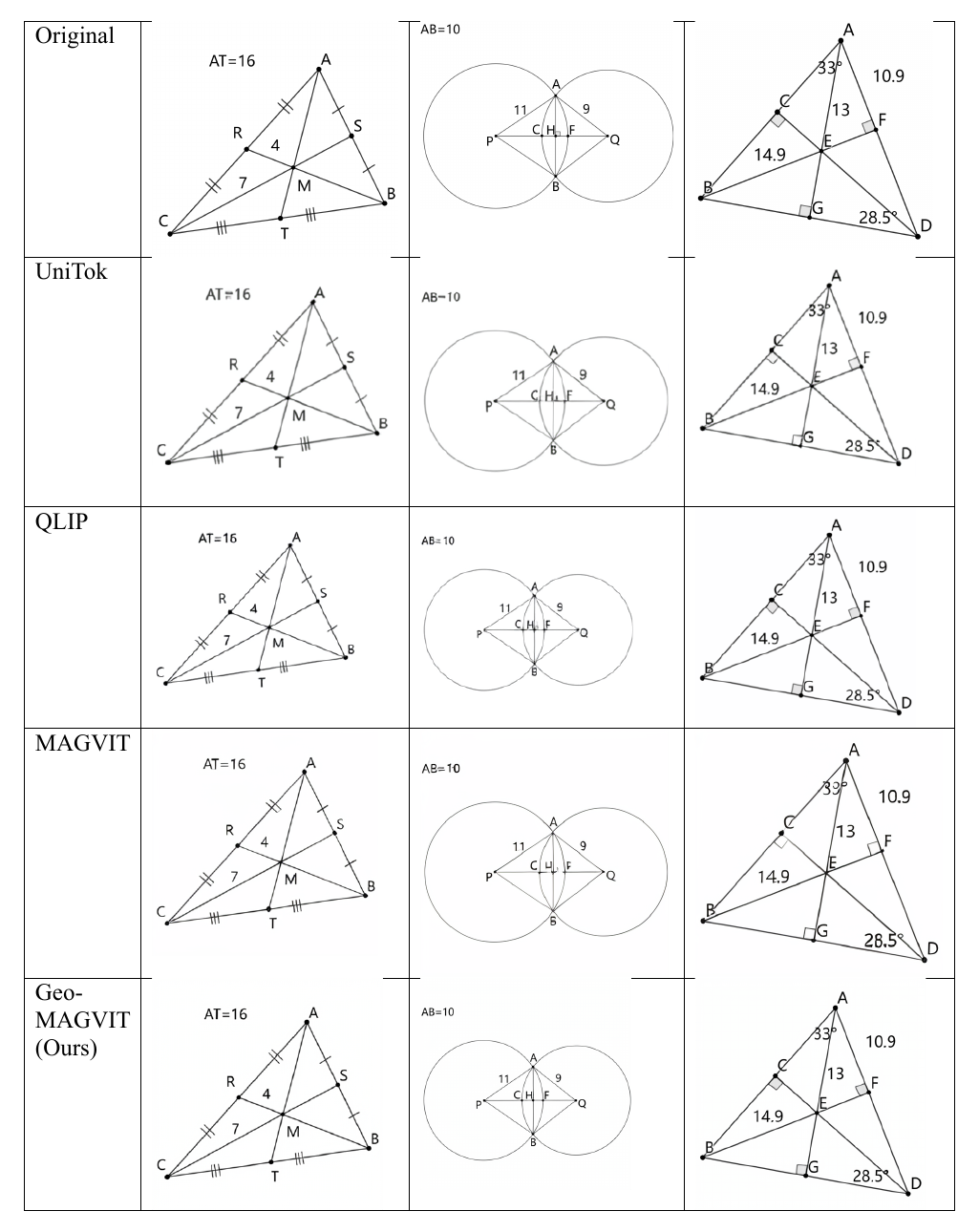}
    \caption{Diagram Reconstruction}
    \label{fig:diagram-rec}
    \Description{Showcase of Diagram Reconstruction}
\end{figure*}

We demonstrate the superior performance of Geo-MAGVIT in diagram reconstruction, as shown in Figure~\ref{fig:diagram-rec}.

\section{Examples for Text to Diagram}
\begin{figure*}[htbp]
    \centering
    \includegraphics[width=0.9\linewidth]{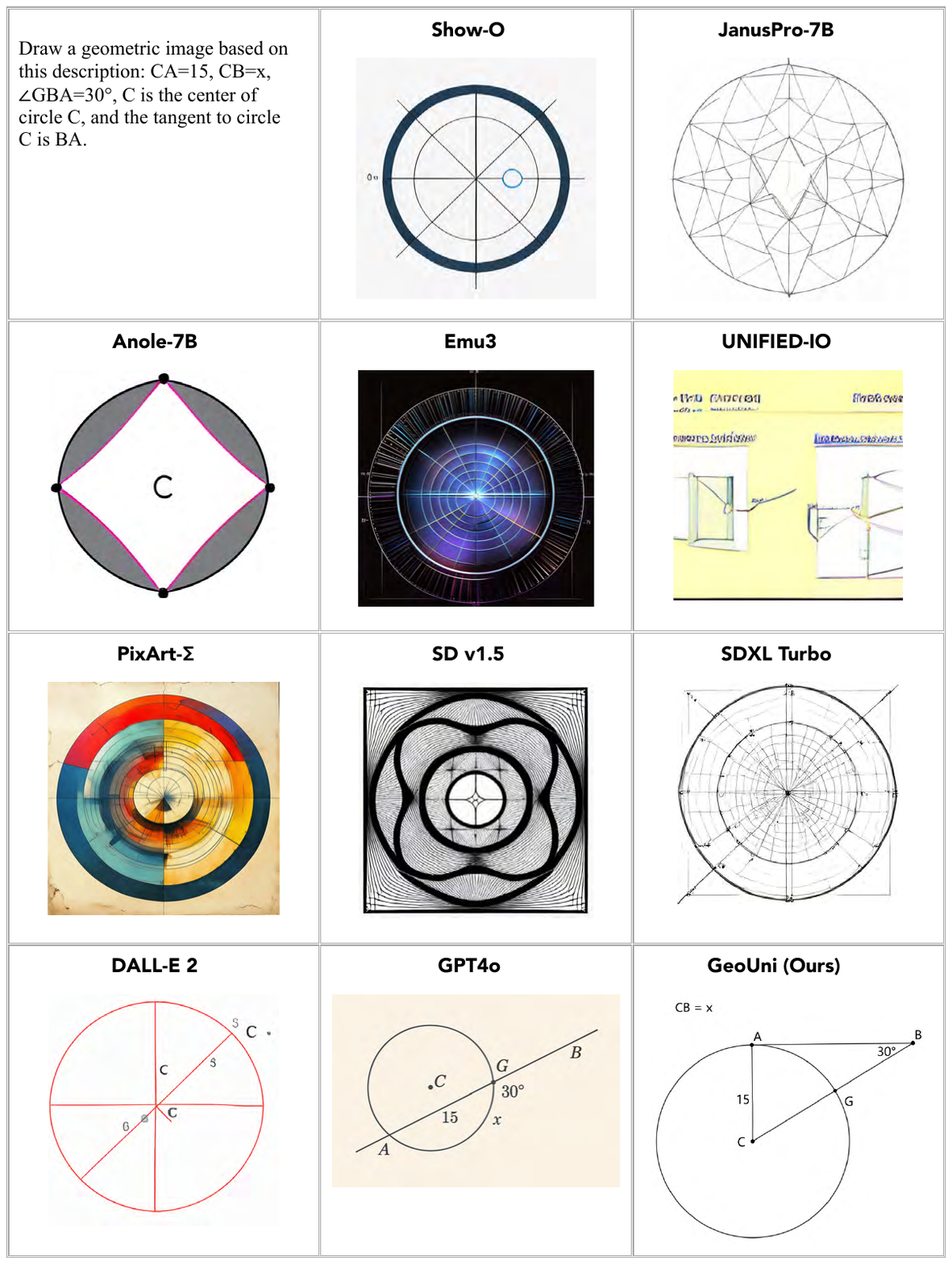}
    \caption{Text-To-Diagram-Showcase-1}
    \label{fig:t2i-1}
    \Description{Showcase of T2I}
\end{figure*}

\begin{figure*}[htbp]
    \centering
    \includegraphics[width=0.9\linewidth]{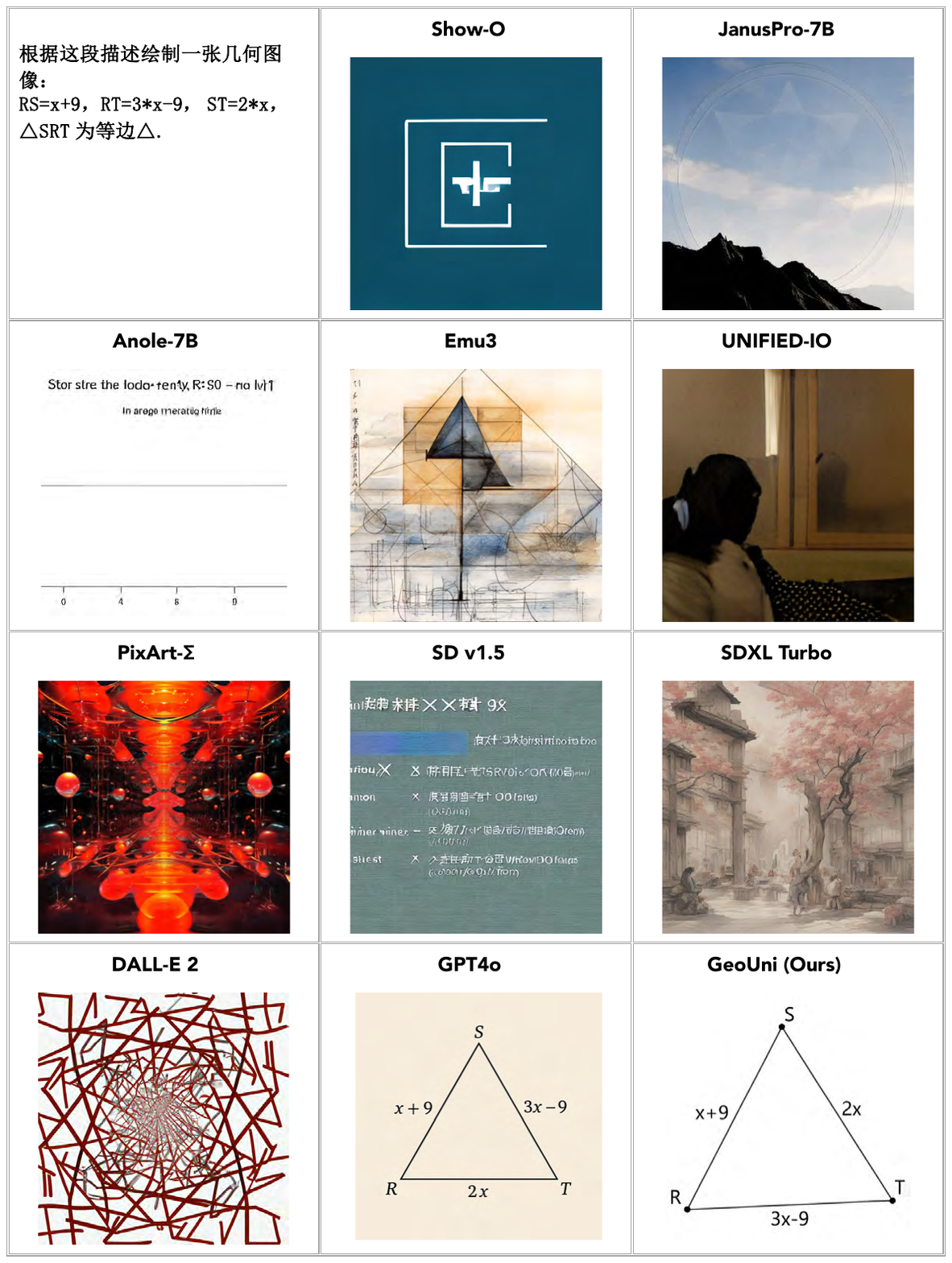}
    \caption{Text-To-Diagram-Showcase-2}
    \label{fig:t2i-2}
    \Description{Showcase of T2I}
\end{figure*}

\begin{figure*}[htbp]
    \centering
    \includegraphics[width=0.9\linewidth]{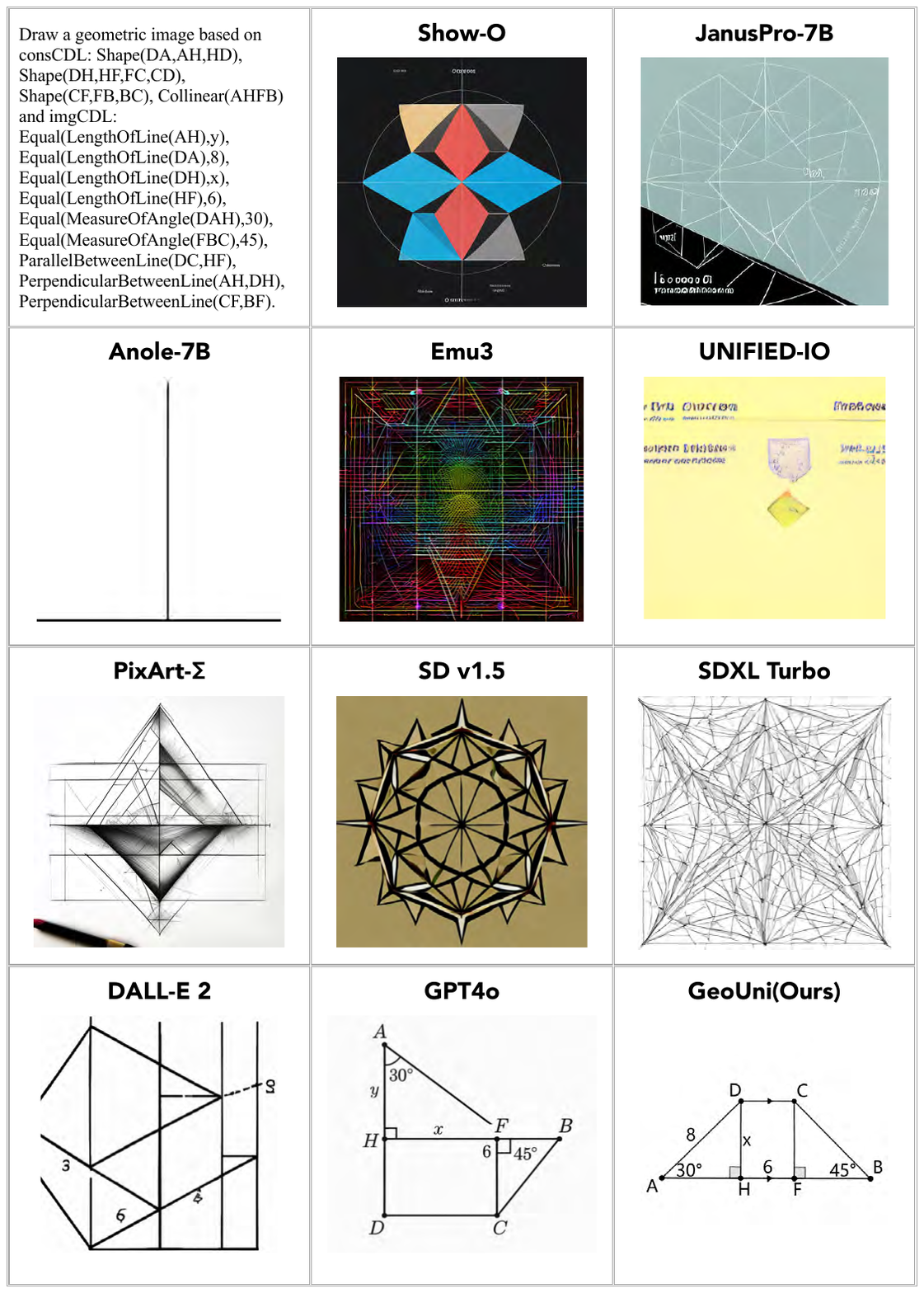}
    \caption{Text-To-Diagram-Showcase-3}
    \label{fig:t2i-3}
    \Description{Showcase of T2I}
\end{figure*}

\begin{figure*}[htbp]
    \centering
    \includegraphics[width=0.9\linewidth]{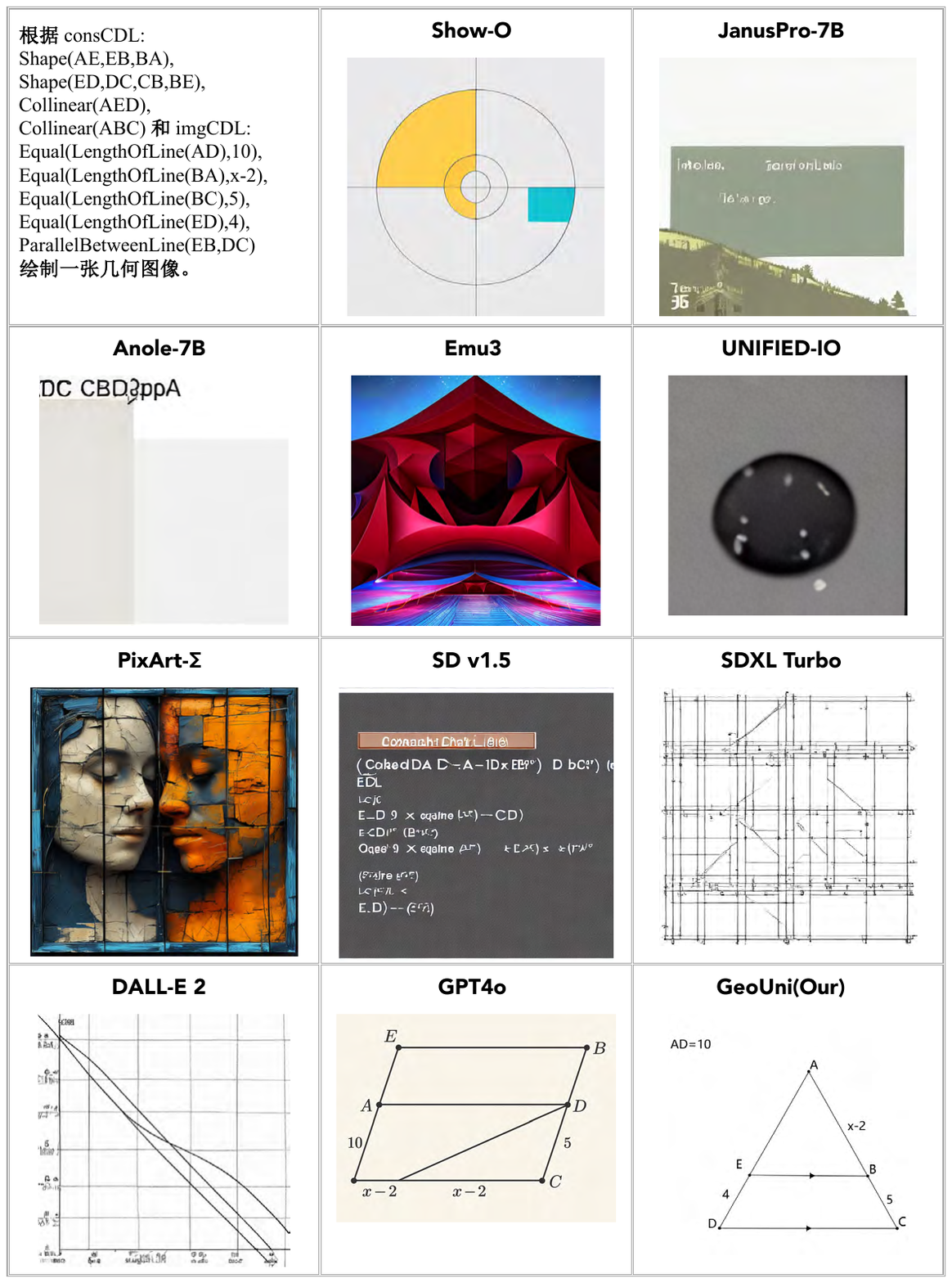}
    \caption{Text-To-Diagram-Showcase-4}
    \label{fig:t2i-4}
    \Description{Showcase of T2I}
\end{figure*}

\begin{figure*}[htbp]
    \centering
    \includegraphics[width=0.9\linewidth]{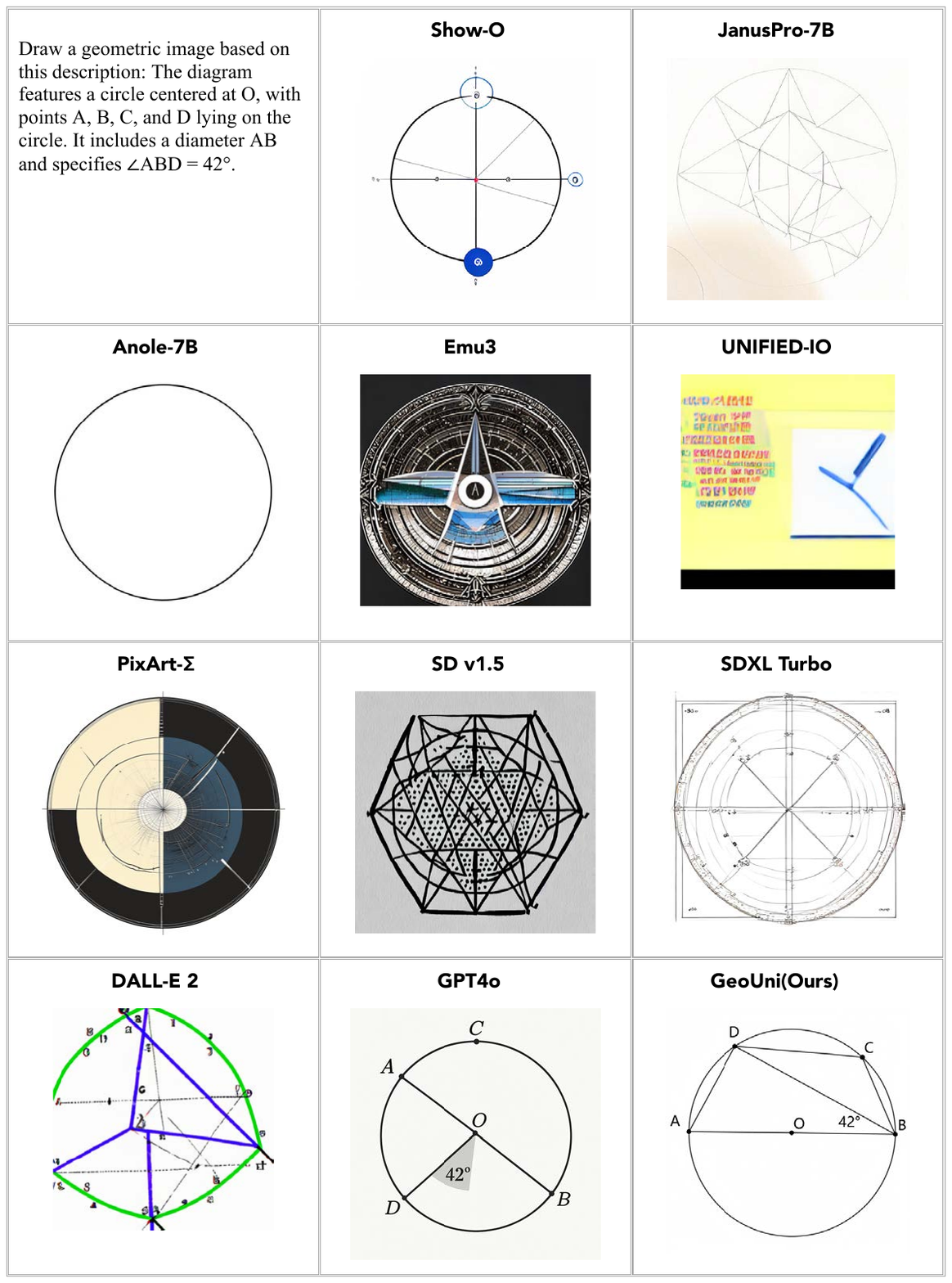}
    \caption{Text-To-Diagram-Showcase-5}
    \label{fig:t2i-5}
    \Description{Showcase of T2I}
\end{figure*}

\begin{figure*}[htbp]
    \centering
    \includegraphics[width=0.9\linewidth]{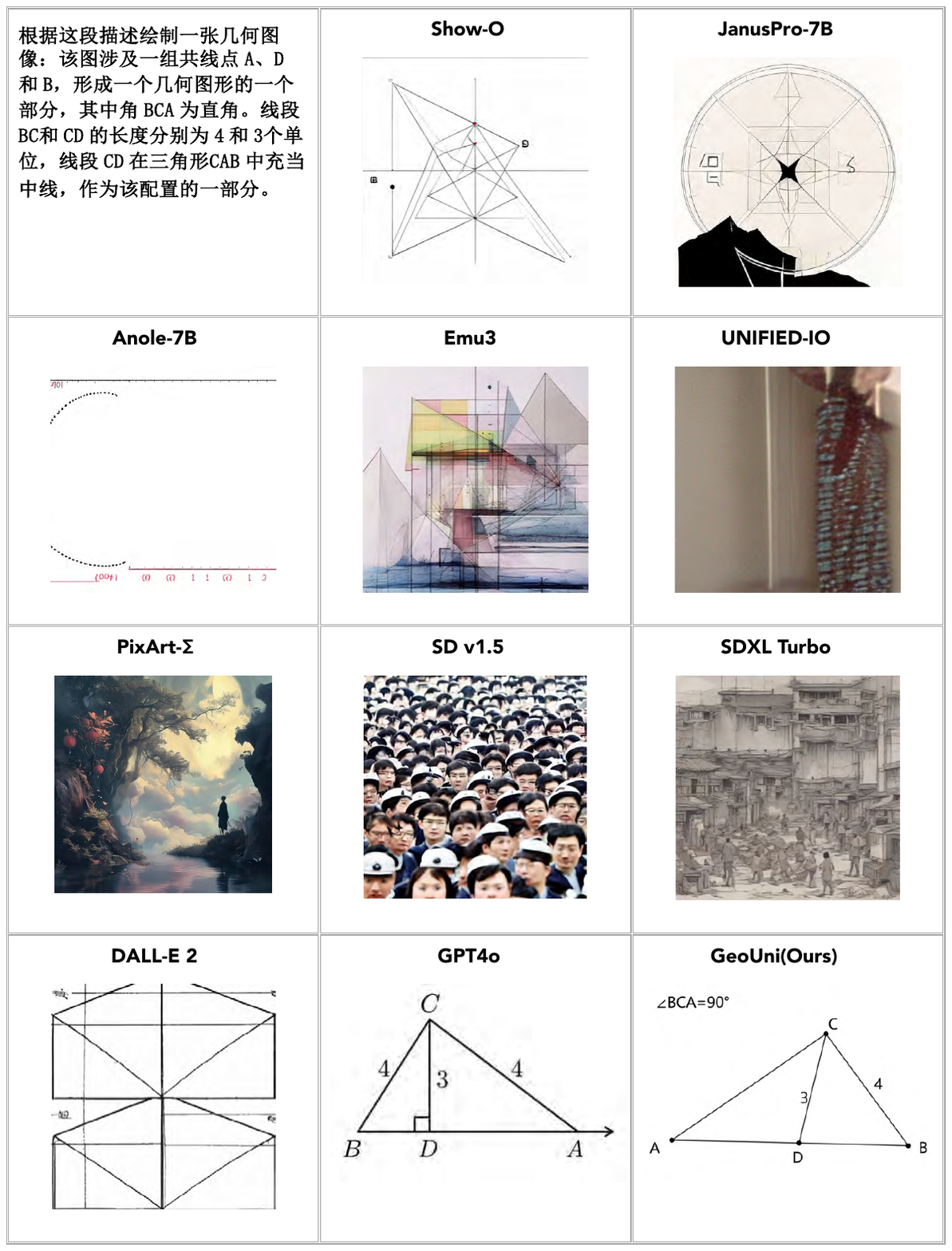}
    \caption{Text-To-Diagram-Showcase-6}
    \label{fig:t2i-6}
    \Description{Showcase of T2I}
\end{figure*}

We demonstrate the superior performance of GeoUni in Text to Diagram based on different prompt formats, as shown in Figures.~\ref{fig:t2i-1}, \ref{fig:t2i-2}, \ref{fig:t2i-3}, \ref{fig:t2i-4}, \ref{fig:t2i-5}, and \ref{fig:t2i-6}.

\section{Examples for Problem Creation}
\begin{figure*}[htbp]
    \centering
    \includegraphics[width=0.9\linewidth]{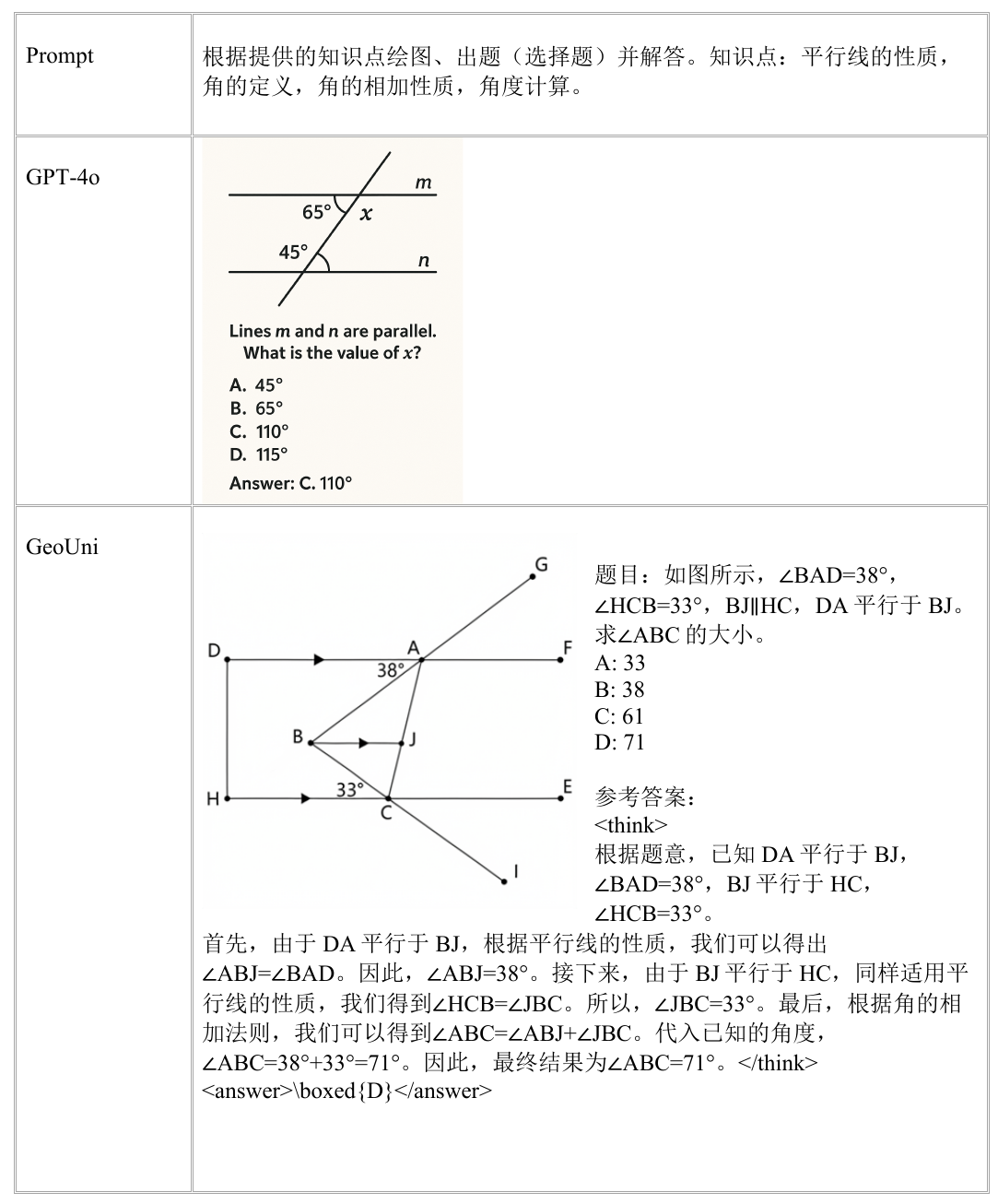}
    \caption{Problem-Creation-Showcase-1}
    \label{fig:problem1}
    \Description{Problem-Creation-Showcase-1}
\end{figure*}

\begin{figure*}[htbp]
    \centering
    \includegraphics[width=0.9\linewidth]{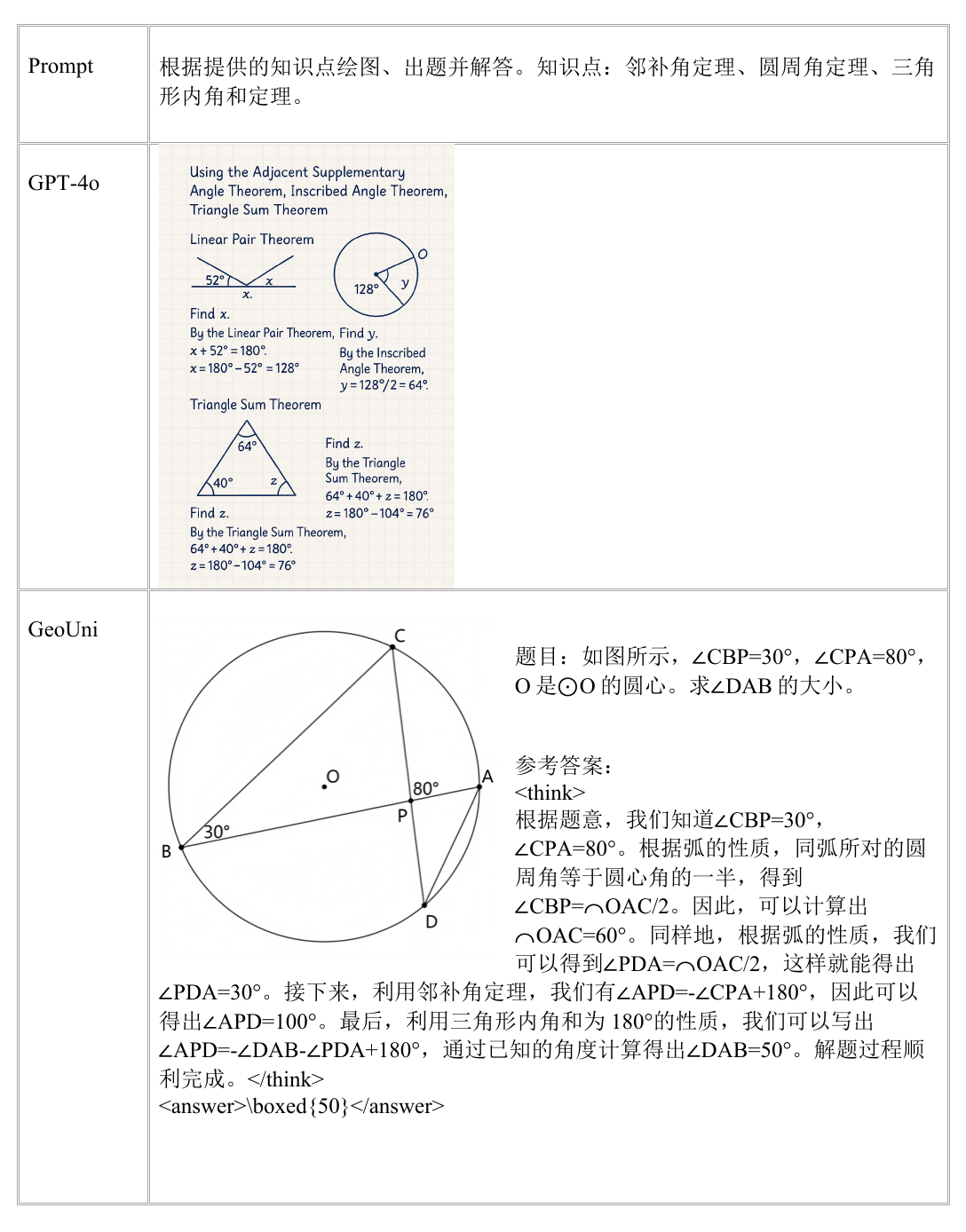}
    \caption{Problem-Creation-Showcase-2}
    \label{fig:problem2}
    \Description{Problem-Creation-Showcase-2}
\end{figure*}

\begin{figure*}[htbp]
    \centering
    \includegraphics[width=0.9\linewidth]{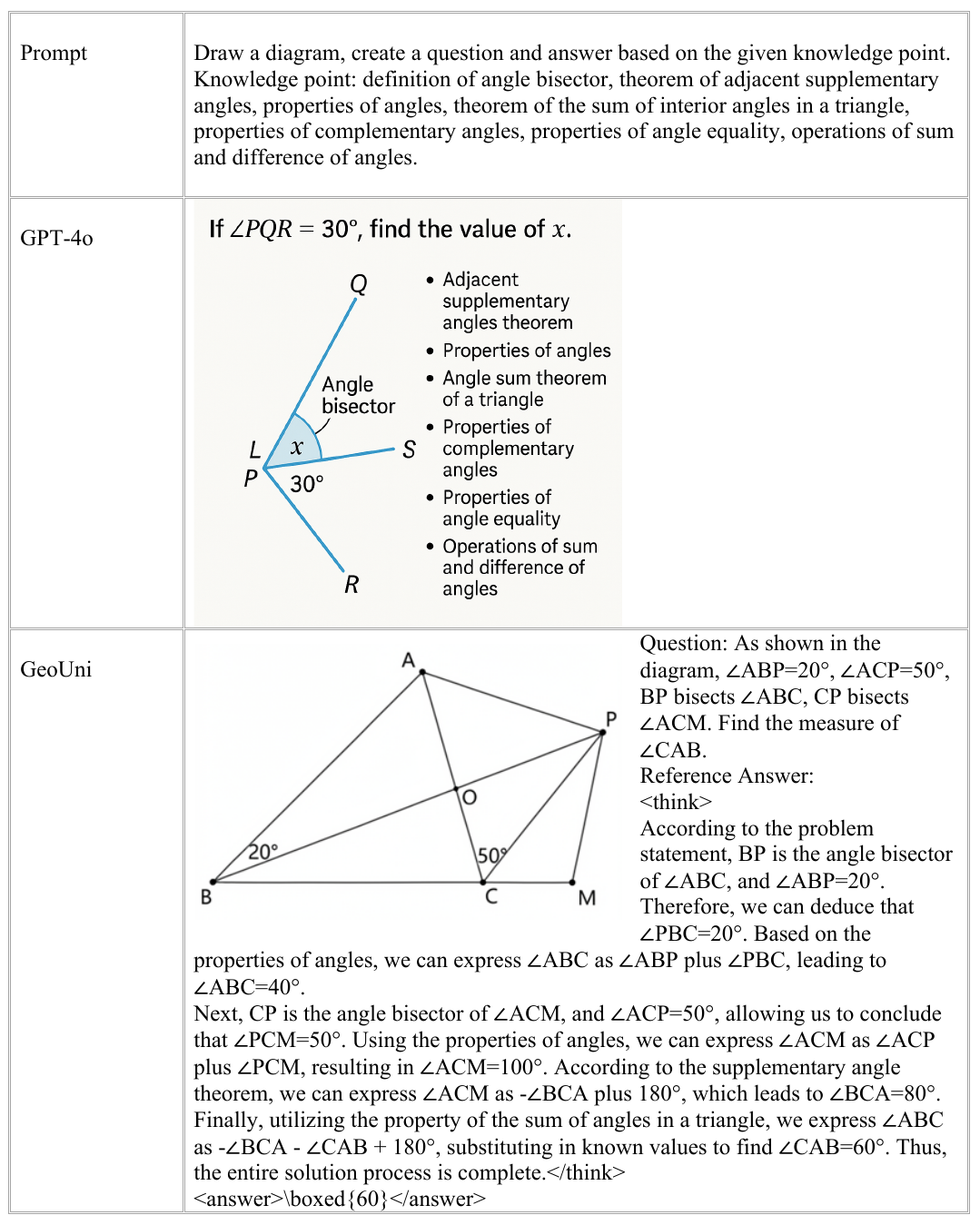}
    \caption{Problem-Creation-Showcase-3}
    \label{fig:problem3}
    \Description{Problem-Creation-Showcase-3}
\end{figure*}

We demonstrate the superior performance of GeoUni in Problem Creation based on English and Chinese prompts, compared to GPT-4o, as shown in Figures.~\ref{fig:problem1}, \ref{fig:problem2}, and \ref{fig:problem3}.


\begin{figure*}[htbp]
    \centering
    \includegraphics[width=0.9\linewidth]{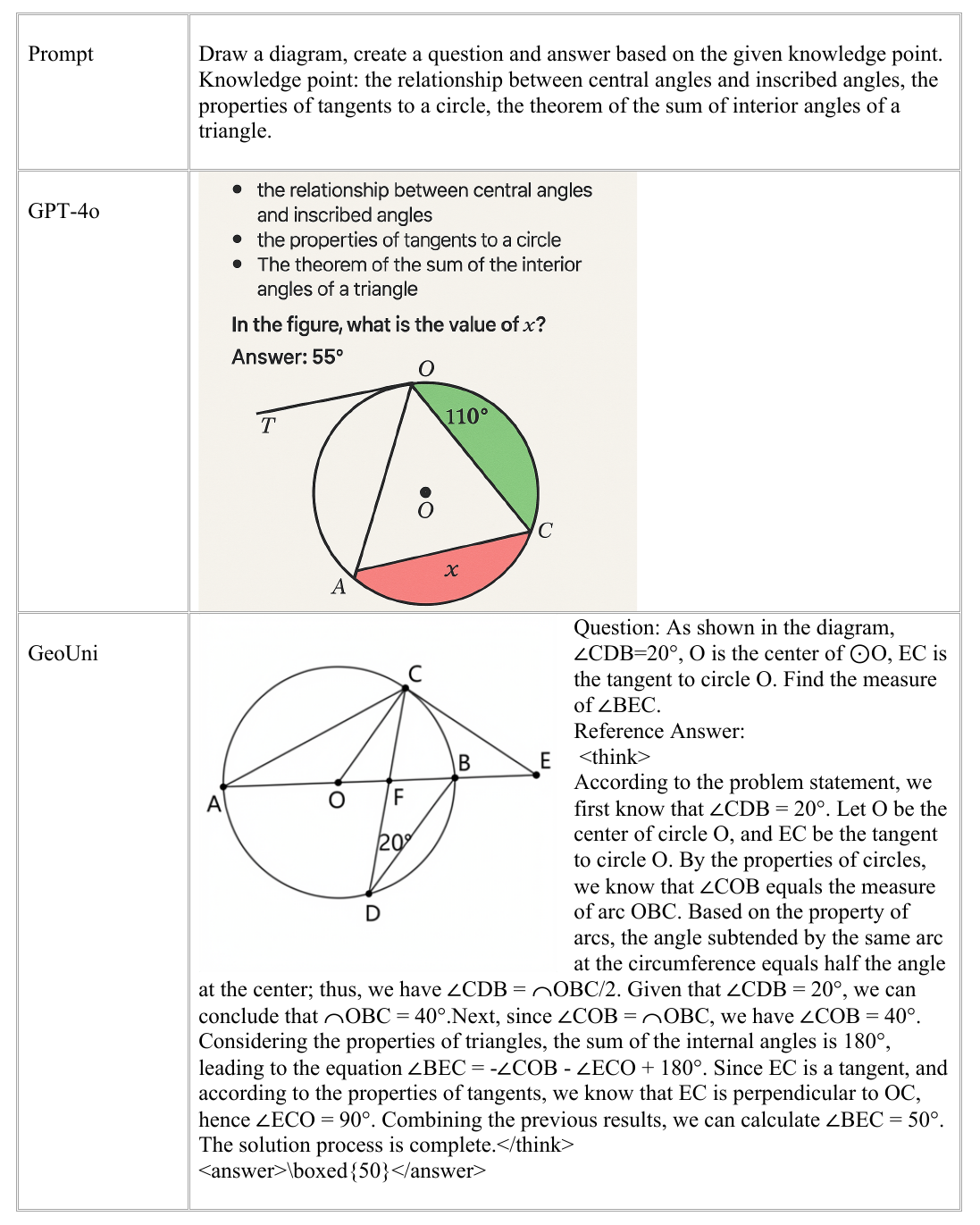}
    \caption{Problem-Creation-Showcase-4}
    \label{fig:proble4m}
    \Description{Problem-Creation-Showcase-4}
\end{figure*}

\end{document}